\documentclass[default]{sn-jnl}

\usepackage{amsmath,amssymb,amsfonts}
\usepackage{graphicx}
\usepackage{textcomp}
\usepackage{mathtools}
\usepackage{comment}
\usepackage{tabularx}
\usepackage{array}
\usepackage{booktabs}
\usepackage{caption}
\usepackage{subcaption}
\usepackage{multirow}

\usepackage{enumitem}
\usepackage{hyperref}

\newcolumntype{P}[1]{>{\centering\arraybackslash}p{#1}}
\newcolumntype{C}[1]{>{\arraybackslash}p{#1}}

\def\BibTeX{{\rm B\kern-.05em{\sc i\kern-.025em b}\kern-.08em
    T\kern-.1667em\lower.7ex\hbox{E}\kern-.125emX}}

\jyear{2022}%

\theoremstyle{thmstyleone}%
\theoremstyle{thmstyletwo}%

\theoremstyle{thmstylethree}%

\begin{document}

\title[COMPASS]{COMPASS: A Formal Framework and Aggregate Dataset for Generalized Surgical Procedure Modeling}


\author*[1]{\fnm{Kay} \sur{Hutchinson}}\email{kch4fk@virginia.edu}

\author[2,3]{\fnm{Ian} \sur{Reyes}}\email{ir6mp@virginia.edu}

\author[1]{\fnm{Zongyu} \sur{Li}}\email{zl7qw@virginia.edu}

\author[1,2]{\fnm{Homa} \sur{Alemzadeh}}\email{ha4d@virginia.edu}

\affil*[1]{\orgdiv{Department of Electrical and Computer Engineering}, \orgname{University of Virginia}, \orgaddress{\city{Charlottesville}, \postcode{22903}, \state{VA}, \country{USA}}}

\affil[2]{\orgdiv{Department of Computer Science}, \orgname{University of Virginia}, \orgaddress{\city{Charlottesville}, \postcode{22903}, \state{VA}, \country{USA}}}

\affil[3]{\orgname{IBM}, \orgaddress{\city{RTP}, \postcode{27709}, \state{NC}, \country{USA}}}


\abstract{
\textbf{Purpose:} We propose a formal framework for the modeling and segmentation of minimally-invasive surgical tasks using a unified set of motion primitives (MPs) to enable more objective labeling and the aggregation of different datasets. 

\textbf{Methods:} We model surgical tasks as finite state machines, representing how the execution of MPs as the basic surgical actions results in the change of surgical context, which characterizes the physical interactions among tools and objects in the surgical environment. We develop methods for labeling surgical context based on video data and for automatic translation of context to MP labels. We then use our framework to create the COntext and Motion Primitive Aggregate Surgical Set (COMPASS), including six dry-lab surgical tasks from three publicly-available datasets (JIGSAWS, DESK, and ROSMA), with kinematic and video data and context and MP labels. 

\textbf{Results:} Our context labeling method achieves near-perfect agreement between consensus labels from crowd-sourcing and expert surgeons. Segmentation of tasks to MPs results in the creation of the COMPASS dataset that nearly triples the amount of data for modeling and analysis and enables the generation of separate transcripts for the left and right tools. 

\textbf{Conclusion:} The proposed framework results in high quality labeling of surgical data based on context and fine-grained MPs. Modeling surgical tasks with MPs enables the aggregation of different datasets and the separate analysis of left and right hands for bimanual coordination assessment. 
Our formal framework and aggregate dataset can support the development of models and algorithms for surgical process analysis, skill assessment, error detection, and autonomy.

}




\maketitle

\section{Introduction}
\label{sec:introduction}
Surgical gestures or surgemes are the building blocks of tasks and represent an important analytical unit for surgical process modeling \cite{lalys2014surgical, neumuth2011modeling}, skill assessment \cite{tao2012sparse, varadarajan2009data}, error detection \cite{yasar2019context, yasar2020real,hutchinson2022analysis,Li2022error, inouye2022assessing}, and autonomy \cite{ginesi2020autonomous}. However, existing datasets and methods for gesture segmentation each use their own set of gesture definitions hindering direct comparisons between them and limiting their compatibility for combined analysis \cite{van2021gesture}. In addition, the descriptive gesture definitions are often subjective, and manual labeling of gestures is tedious and inconsistent. A recent survey of the state-of-the-art research on surgical gesture recognition \cite{van2021gesture} highlighted the need for a common surgical language with defined segmentation boundaries, as well as larger datasets to support comparative analysis and future work in error detection and prediction.

There are many datasets from real surgical tasks, but they only contain video data and are predominantly used for tool and object recognition such as those used for the EndoVis challenge \cite{allan20202018, wagner2021comparative} and CholecT50 \cite{nwoye2022cholectriplet2021}. 
However, datasets with both kinematic and video data from a surgical robot are small and contain only a handful of trials of a few simulated or dry-lab tasks performed by a limited number of subjects. This hinders analysis and the training of machine learning (ML) models especially in the areas of surgical process modeling, gesture recognition, autonomy, and error detection \cite{van2021gesture}. This scarcity of data also means ML models will see subjects, trials, and \textit{tasks} that could be very different from their training set when they are deployed. 

The most commonly used dataset for training and evaluation of different gesture recognition models is JIGSAWS \cite{gao2014jhu} which contains kinematic data, videos, gesture labels, and surgical skill scores for three dry-lab surgical tasks. However, only two of its tasks, Suturing and Needle Passing, are labeled with similar sets of gestures; Knot Tying only shares the same first and last gestures with the other tasks. 
In addition, recent studies have indicated inconsistency and imprecise boundaries in the gesture labels. For example, \cite{van2020multi} made 12 amendments to the gesture labels and \cite{hutchinson2022analysis} identified a significant discrepancy in the annotation of certain gestures that may effect error detection. 
Other recently developed datasets such as DESK and ROSMA either use very different gesture definitions at lower levels of granularity or do not provide gesture labels.
Furthermore, all subjects in JIGSAWS have been right handed and gesture labels cannot be divided into actions performed by the left and right hands separately to allow for bimanual coordination assessments and analysis such as in \cite{boehm2021online}.

Recent works in gesture recognition have each defined their own sets of gestures for their own datasets \cite{dipietro2019segmenting, goldbraikh2022using, menegozzo2019surgical, gonzalez2020desk, de2021first} with limited overlap between gestures. 
Action triplets \cite{meli2021unsupervised, li2022sirnet, nwoye2022rendezvous} have also been proposed for surgical activity recognition by considering the interactions among tools and objects, but they have mainly focused on only video data from real surgery. Kinematic data is very valuable for safety analysis~\cite{yasar2020real, Li2022error, hutchinson2022analysis}, improved recognition accuracy using multi-modal analysis~\cite{qin2020temporal, qin2020davincinet}, or when video data is not available or noisy~\cite{yasar2019context} due to smoke or occlusions in the surgical environment.
Another recent work modeled surgical processes using statecharts with surgemes and triggers \cite{falezza2021modeling}. Section \ref{sect:related_work} presents a detailed summary of related work on gesture and action definitions and datasets, where the granularity of the actions 
can vary from the sub-gesture to the task level (see Figure~\ref{fig:hierarchy}). 

A significant challenge is that there is still no formal framework that defines a standard set of surgical actions and their relations to gestures and tasks, which would enable direct comparisons between these works, their datasets, and models of their tasks. 

Our contributions are as follows:
\begin{itemize}
\item We propose a novel formal framework for modeling surgical tasks with finite state machines using a standardized set of motion primitives whose execution results in changes in important state variables that make up the surgical context. In this framework, surgical context characterizes the physical interactions among surgical objects and instruments, and motion primitives represent the basic surgical actions across different surgical tasks and procedures.
\item We develop a method for labeling surgical context based on video data that achieves near-perfect agreement between crowd-sourced labels and expert surgeon labels, higher agreement among annotators than existing gesture definitions, and such that the context labels can be automatically translated into motion primitive labels. 
\item We apply our framework and labeling method to create an aggregate dataset, called COMPASS (COntext and Motion Primitive Aggregate Surgical Set), consisting of kinematic and video data as well as context and motion primitive labels for a total of six dry-lab tasks from the JIGSAWS \cite{gao2014jhu}, DESK \cite{gonzalez2020desk}, and ROSMA \cite{rivas2021surgical} datasets. 

\end{itemize}

The tools for labeling surgical context based on video data and automated translation of context to motion primitive labels as well as the aggregated dataset with context and motion primitive labels are made publicly available at \url{https://github.com/UVA-DSA/COMPASS} to facilitate further research and collaboration in this area.

\section{Methods}

Our framework models surgical procedures as a language with a grammar dictating how motion primitives are combined to perform gestures and tasks, thus bridging the gap between semantic-less motions described by \cite{neumuth2011modeling} and \cite{lalys2014surgical}, and intent-based gestures described by \cite{lin2010structure}. In our framework, we formally define surgical motion primitives, how they relate to surgical context and task progress, and how they can be combined to perform tasks. To accomplish this, we develop methods for the objective labeling of surgical context and translation of context labels to motion primitive labels, and apply them to three publicly available datasets to create the aggregated COMPASS dataset. 

\subsection{Modeling Framework}\label{sec:Framework}

\subsubsection{Surgical Hierarchy}
\begin{figure}
    \centering
    \includegraphics[trim = 0in 2.8in 5in 0in, width=0.75\textwidth]{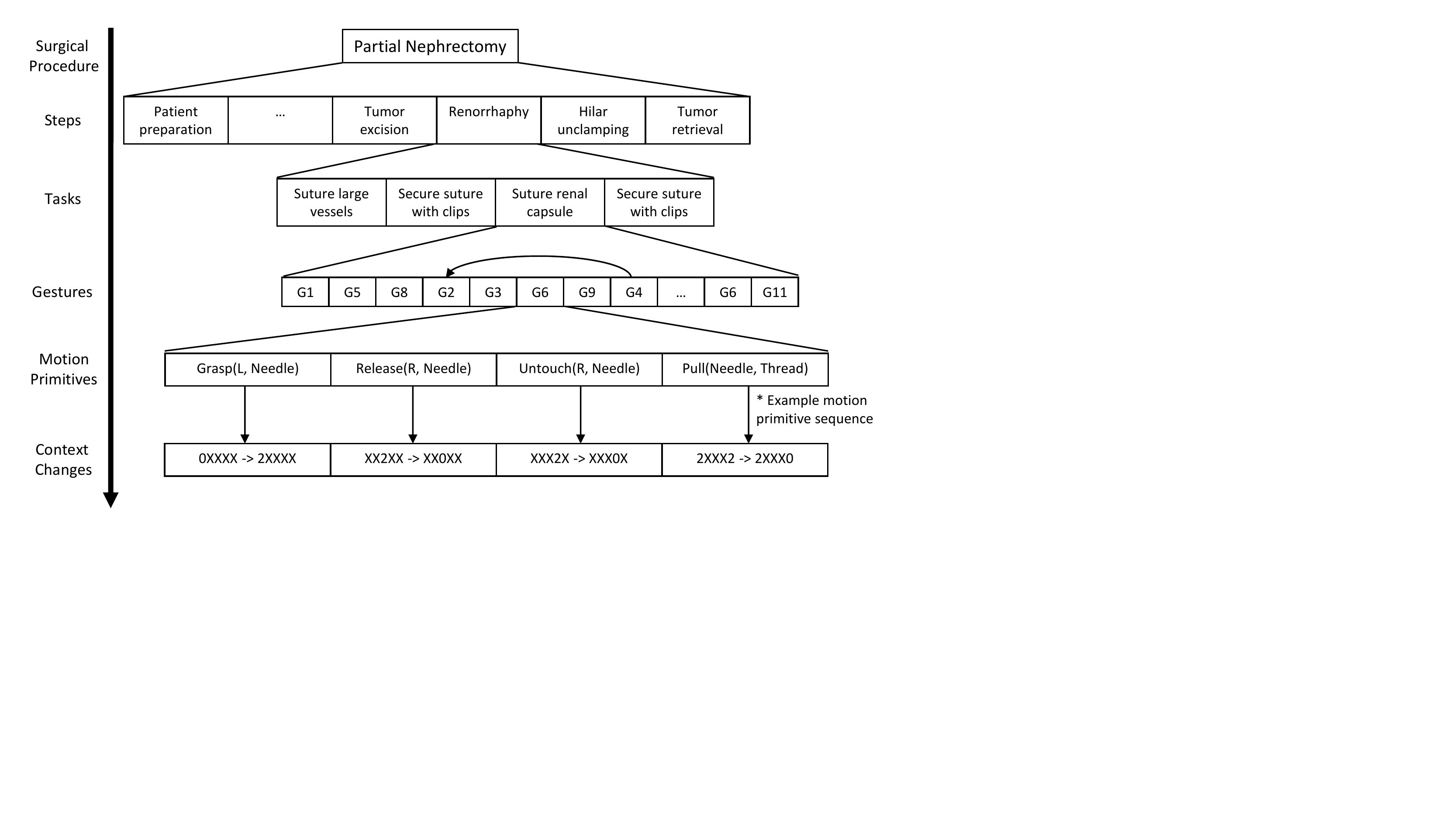}
    \caption{Surgical Hierarchy. Adapted from \cite{hutchinson2022analysis}}
    \label{fig:hierarchy}
    \vspace{-0.5em}
\end{figure}

Surgical procedures follow the hierarchy of levels defined in \cite{neumuth2011modeling} which provides context \cite{yasar2020real} for actions during the procedure, as shown in Figure \ref{fig:hierarchy}. A surgical \textbf{operation} can involve multiple \textbf{procedures} which are divided into \textbf{steps}. Each \textbf{step} is subdivided into \textbf{tasks} comprised of \textbf{gestures} (also called sub-tasks or surgemes). These \textbf{gestures} are made of basic \textbf{motion primitives} such as moving an instrument or closing the graspers, which effect changes in important states that comprise the overall surgical \textbf{context}. 

\begin{figure*}[t!]
    \centering
    \begin{minipage}{0.49\textwidth}
        \begin{subfigure}{\textwidth}
            \centering
            \includegraphics[trim = 0.1in 2.5in 9in 0.25in, clip, 
            width=\textwidth]{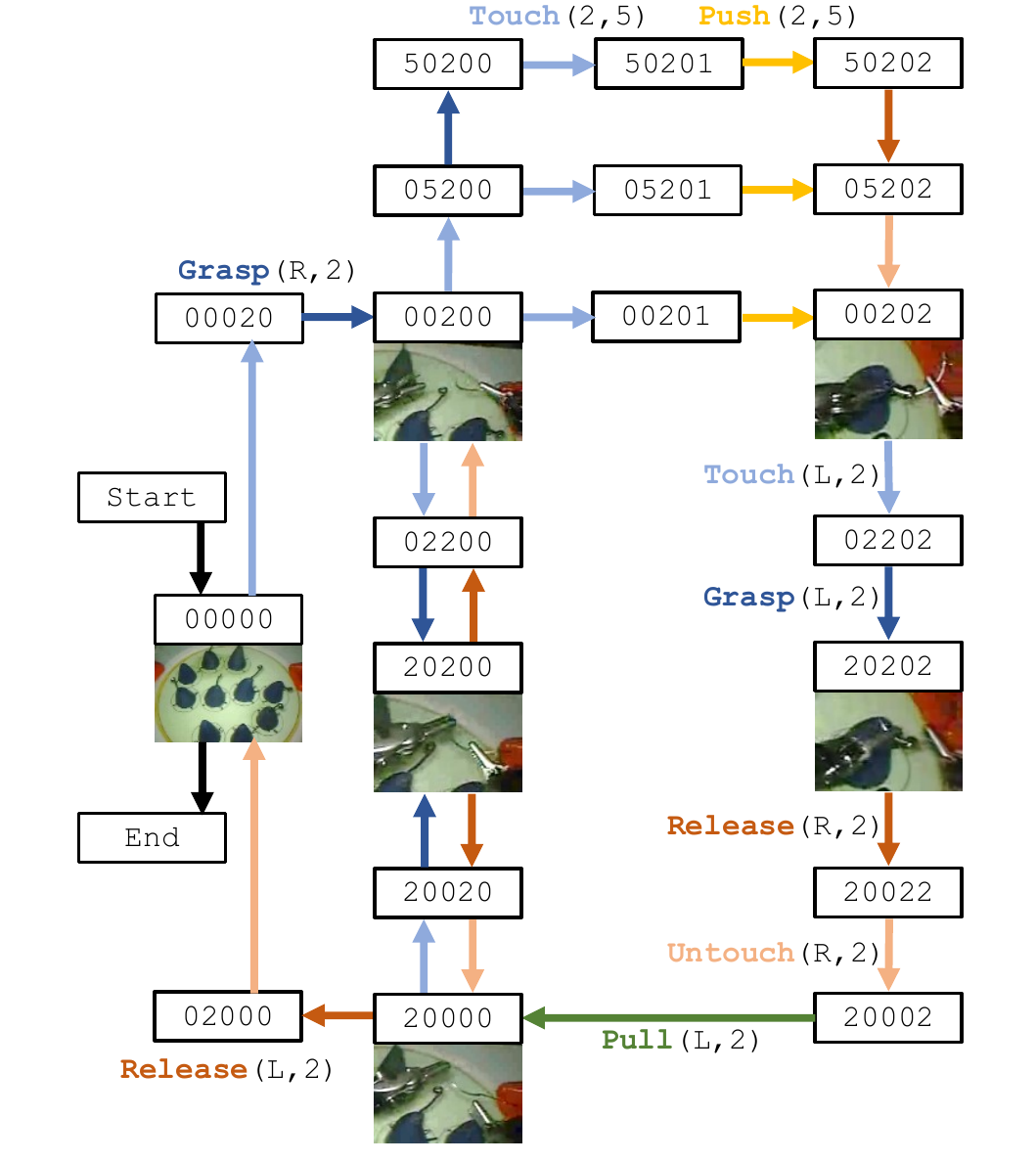}
            \caption{}
            \label{fig:NP_graph}
        \end{subfigure}
    \end{minipage}
    \begin{minipage}{0.49\textwidth}
        \begin{subfigure}{\textwidth}
            \centering
            \includegraphics[trim = 4.4in 5in 4in 0.25in, clip, 
            width=\textwidth]{Figures/NP_graph.pdf}
            \caption{}
            \label{fig:context}
        \end{subfigure}
        
        \vfill
        \begin{subfigure}{\textwidth}
            \centering
            \includegraphics[trim = 4.4in 2.5in 4in 3in, clip, 
            width=\textwidth]{Figures/NP_graph.pdf}
            \caption{}
            \label{fig:MPsandgestures}
        \end{subfigure}
    \end{minipage}
    \hfill 
    \vspace{-0.5em}
    \caption{Modeling the Needle Passing Task: \ref{fig:NP_graph}) The finite state machine model representing the ideal performance of a Needle Passing trial with context and MPs. \ref{fig:context}) An example of a motion primitive in Needle Passing showing the state variables, objects, and needle states. \ref{fig:MPsandgestures}) Example of alignment between MPs and gestures in a Needle Passing trial that also shows the discrepancy in the G3 boundary as noted by \cite{hutchinson2022analysis} where the 'Push' MP is not part of G3. From \cite{gao2014jhu}, G2: positioning needle, G3: pushing needle through tissue, G4: transferring needle from left to right, G6: pulling suture with left hand. Figure best viewed in color.} 
    \label{fig:NP_model}
    \vspace{-1em}
\end{figure*}

\subsubsection{Surgical Context}
A surgical environment (either during dry-lab or real surgical procedures) can be modeled by a set of state variables that characterize the status and interactions among surgical instruments (e.g., graspers, scissors, electro-cautery) and objects (e.g., needles, threads, blocks, balls, sleeves, rings) or anatomical structures (e.g., organs, tissues, tumors) at a given time in the physical environment. 
Changes in the surgical context happen as the result of performing a set of basic motion primitives by the robot (either controlled by the surgeon operator or autonomously). We model each surgical task as a finite state machine with the states representing the surgical context and the transitions representing the motion primitives. Figure \ref{fig:NP_graph} shows an example of the finite state machine model for the Needle Passing task. This new representation of surgical tasks enables the incorporation of surgical context into surgical procedure modeling which is missing from the previously proposed models such as grammar graphs and Hidden Markov models \cite{ahmidi2017dataset} where lower level actions are obscured by hidden states.

We define the surgical context using two sets of variables that can be observed or measured using kinematic and/or video data from a surgical scene: (i) general state variables relating to the contact and hold interactions between the tools and objects in the environment, and (ii) task-specific state variables describing the states of objects critical to the current task. We also define independent state variables for the left and right tools to enable the generation of separate label sets to support side-specific skill assessment, hand coordination analysis, and improved MP recognition. 
There are four general state variables as shown in Figure \ref{fig:context}. An additional task-specific state variable is appended to the right of the general context variables to describe progress in the task. For Suturing and Needle Passing, the needle, if held, can be ``not touching", ``touching", or ``in" the fabric or ring. For Knot Tying, the thread can be ``wrapped" around the opposite grasper, in a ``loose" knot, or in a ``tight" knot. For Peg Transfer and Post and Sleeve, the block can be ``on" or ``off" the peg. For Pea on a Peg, the pea, if held, can be ``in the cup", ``stuck to other peas", ``not stuck to other peas", or ``on the peg". 
For example, in Figure \ref{fig:context}, the state 50202 indicates that the left grasper is holding a ring, the right grasper is holding the needle, and the needle is in the ring.

\subsubsection{Motion Primitives}
We define a unified set of six modular and programmable surgical motion primitives (MPs) to model the basic surgical actions that lead to changes in the physical context. As shown in Equation \ref{MP_eqn}, each MP is characterized by its type (e.g., Grasp), the specific tool which is used (e.g., left grasper), the object with which the tool interacts (e.g., block), and a set of constraints that define the functional (e.g., differential equations characterizing typical trajectory \cite{ginesi2019dmp++, ginesi2020autonomous, ginesi2021overcoming}) and safety requirements (e.g., virtual fixtures and no-go zones \cite{bowyer2013active, sutherland2015robotics, yasar2019context}) for the execution of the MPs: 
\vspace{-0.5em}
\begin{equation}
    MP(tool, object, constraints)
    \label{MP_eqn}
\end{equation}

In this framework, tools and objects are considered classes as in object-oriented programming and can have attributes such as the specific type of tool and current position. Also, the MPs can be further decomposed into the fundamental transformations of move/translate, rotate, and open/close graspers which characterize the low level kinematic commands for the programming and execution of motions on a robot, but we do not examine that level here.  

Our MPs are similar to the recently proposed action triplets in \cite{nwoye2022rendezvous} for surgical activity recognition based on video data in real surgical tasks. But we focus on developing and applying standardized labels to dry-lab datasets with \textit{both kinematic and video data} to enable comparative analyses between datasets and tasks. Kinematic data can support analysis for safety and skill, and be used to develop dynamic motion primitives (DMPs) \cite{ginesi2019dmp++}. The COMPASS framework can be extended to real surgical procedures by adding additional tool-specific verbs similar to those proposed in \cite{nwoye2022rendezvous} (e.g., "Cut" for scissors).
Segmenting tasks into MPs allows the separation of actions performed by the left and right hands and the generation of separate sets of labels which can support more detailed skill assessment, analysis of bimanual coordination~\cite{boehm2021online}, and surgical automation~\cite{van2021gesture}. 
To generate separate left and right label sets, MPs performed by each hand or arm of the robot are split into new transcripts and the 'Idle' MP is used to fill the gaps created by the separation so that every kinematic sample has a label.

Table \ref{tab:motion_primitives_general} shows the set of MPs and corresponding changes to surgical context applicable to all tasks. Table \ref{tab:motion_primitives_task_specific} shows the sets of MPs and corresponding changes to surgical context applied to specific dry-lab tasks in our aggregated dataset. In this work, we only focus on dry-lab tasks where the tools are graspers. We also do not model or analyze the MP-specific functional and safety constraints since we only focus on recognition and not automation or monitoring.

\begin{table}[t!]
    \centering
    \caption{General motion primitives for changes in context: `L' and `R' represent the left and right graspers as tools, `a' is a generic object as listed in Fig \ref{fig:context}, and `X' can be any value.}
    \label{tab:motion_primitives_general}
    \begin{tabular}{l P{0.3\linewidth}}
    \toprule
    Motion Primitive & Context Change \\ \midrule
    Touch(L, a) & X0XX $\rightarrow$ XaXX \\
    Touch(R, a) & XXX0 $\rightarrow$ XXXa \\
    Grasp(L, a) & 0aXX $\rightarrow$ aXXX \\ 
    Grasp(R, a) & XX0a $\rightarrow$ XXaX \\ 
    Release(L, a) & aXXX $\rightarrow$ 0aXX \\ 
    Release(R, a) & XXaX $\rightarrow$ XX0a \\ 
    Untouch(L, a) & XaXX $\rightarrow$ X0XX \\ 
    Untouch(R, a) & XXXa $\rightarrow$ XXX0 \\ 
    \bottomrule
\end{tabular}
\vspace{-1em}
\end{table}

\begin{table}[t!]
\centering
\caption{Task-specific motion primitives for changes in context: `L' and `R' represent the left and right graspers as tools, objects are encoded as in Fig \ref{fig:context}, `b' is a value greater than 0, and `X' can be any value.}
\label{tab:motion_primitives_task_specific}
\begin{tabular}{P{0.005\linewidth} l P{0.3\linewidth}}
\toprule
& Motion Primitive & Context Change \\ \midrule

\multicolumn{3}{l}{Suturing/Needle Passing} \\
& Touch(2, 4/5) & 2XXX0 $\rightarrow$ 2XXX1 \\ 
& Touch(2, 4/5) & XX2X0 $\rightarrow$ XX2X1 \\ 
& Push(2, 4/5) & 2XXX1 $\rightarrow$ 2XXX2 \\ 
& Push(2, 4/5) & XX2X1 $\rightarrow$ XX2X2 \\ 
& Pull(2, 3) & 2XXX2 $\rightarrow$ 2XXX0 \\ 
& Pull(2, 3) & XX2X2 $\rightarrow$ XX2X0 \\ \midrule

\multicolumn{3}{l}{Knot Tying} \\
& Pull(L, 3) & 3XXX0 $\leftrightarrow$ 3XXX1 \\ 
& Pull(R, 3) & XX3X0 $\leftrightarrow$ XX3X1 \\ 
& Pull(L, 3) Pull(R, 3) & 3X3X1 $\rightarrow$ 3X3X2 \\ 
& Pull(L, 3) Pull(R, 3) & 3X3X2 $\rightarrow$ 3X3X3 \\ \midrule

\multicolumn{3}{l}{Peg Transfer and Post and Sleeve} \\
& Touch(1, Post) & XXXX0 $\rightarrow$ XXXX1 \\ 
& Untouch(1, Post) & XXXX1 $\rightarrow$ XXXX0 \\ \midrule

\multicolumn{3}{l}{Pea on a Peg} \\
& Grasp(L, 1) & 0XXX0 $\rightarrow$ 1XXX1 \\ 
& Grasp(R, 1) & XX0X0 $\rightarrow$ XX1X1 \\ 
& Pull(L, 1) & 1XXX1 $\rightarrow$ 1XXX2 \\ 
& Pull(R, 1) & XX1X1 $\rightarrow$ XX1X2 \\ 
& Pull(L, 1) & 1XXX1 $\rightarrow$ 1XXX3 \\ 
& Pull(R, 1) & XX1X1 $\rightarrow$ XX1X3 \\ 
& Touch(1, 1) & XXXX3 $\rightarrow$ XXXX2 \\
& Untouch(1, 1) & XXXX2 $\rightarrow$ XXXX3 \\ 
& Touch(1, Peg) & XXXX3 $\rightarrow$ XXXX4 \\ 
& Untouch(1, Peg) & XXXX4 $\rightarrow$ XXXX3 \\ 
& Release(L, 1) & 1XXXb $\rightarrow$ 0XXX0 \\ 
& Release(R, 1) & XX1Xb $\rightarrow$ XX0X0 \\ 
& Push(L, 1) & 1XXX2 $\rightarrow$ 1XXX1 \\ 
& Push(R, 1) & XX1X2 $\rightarrow$ XX1X1 \\ 
\bottomrule
\end{tabular}
\end{table}

The definition of MPs based on the changes in the surgical context could enable the translation of 
context and MPs to existing gesture labels and facilitate aggregation of different datasets labeled with different gesture definitions. Figure \ref{fig:MPsandgestures} shows an example alignment between MPs and gestures in a Needle Passing trial from the JIGSAWS dataset. We will discuss the labeling of context and automatic translation from context labels to MPs in the next section. However, the automated translation from context and MP labels to existing gesture definitions is complicated, because of the possibility of executional and procedural errors as defined in \cite{hutchinson2022analysis}, and is, thus, beyond the scope of this paper. 

\subsection{Labeling of Context and Motion Primitives}
\subsubsection{Context Labeling}
Gesture recognition models using supervised learning require a large number of annotated video sequences \cite{park2012crowdsourcing}. However, manual labeling of gestures is time consuming and subjective which can lead to labeling errors \cite{van2020multi}. To address this, we have developed a tool for manual annotation of the surgical context (states of the objects and instruments) based on video data and used it to label all trials in six tasks from the JIGSAWS, DESK, and ROSMA datasets.

\begin{figure}[t!]
    \centering
    \includegraphics[trim = 0in 0in 0in 0in, clip, width=0.45\textwidth]{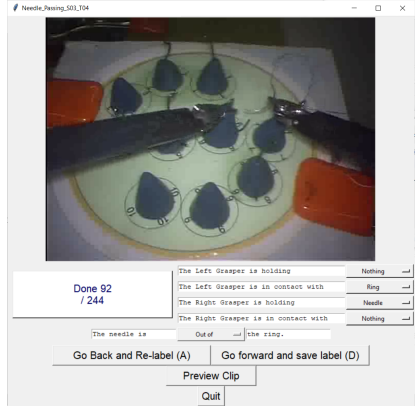} 
    \caption{App for Context Labeling based on Video Data}
    \label{fig:app}
    \vspace{-1em}
\end{figure}

Labeling video data for surgical context provides a more objective way of recognizing gestures and thus can lead to a higher level of agreement among annotators. Also, as noted in \cite{Kitaguchi2021Artificial}, labels for surgical workflow require guidance from surgeons while annotations for surgical instruments do not. Since context labels document the objects held by or in contact with the left and right graspers, they rely less on surgical knowledge than gestures which require anticipating the next actions in a task to mark when a gesture has ended. 
Figure \ref{fig:app} shows a snapshot of the tool for manual labeling of context based on video data. The annotators indicate the value of different state variables for frames in the video data and have the option to copy over the same values of state variables for similar frames until a change in context is observed. This differs from other labeling methods where annotators mark the start and end of each segment and assign it a label.

\subsubsection{Context to Motion Primitive Translation}
\begin{figure}[t!]
    \centering
    \includegraphics[trim = 0in 4.5in 4in 0in, clip, width=0.75\textwidth]{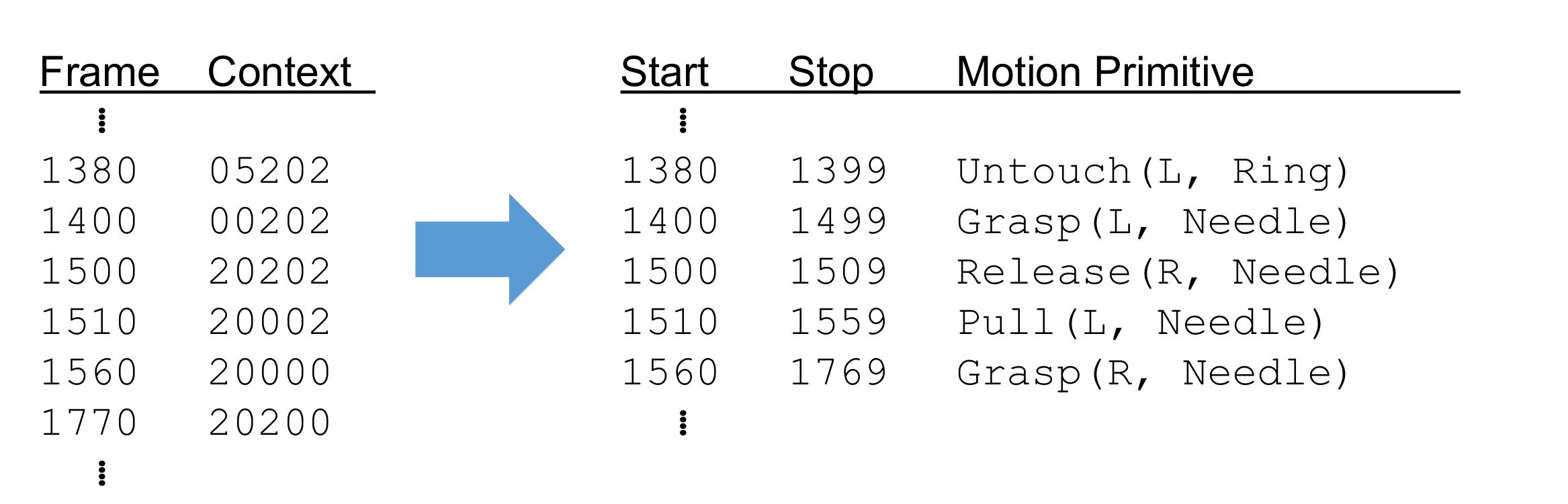}
    \caption{Example sequence of context translated into motion primitives.}
    \label{fig:translation}
    \vspace{-1em}
\end{figure}

Context to MP translation allows us to leverage high quality context labels in creating surgical workflow annotations and aggregating different surgical datasets.
The context labels are translated automatically into MP labels using the finite state machine (FSM) models for each task. In these models, the states are specific contexts and the transitions between states are MPs. Given an input sequence of context labels, the corresponding sequence of motion primitive labels are generated based on the transitions described in Tables \ref{tab:motion_primitives_general} and \ref{tab:motion_primitives_task_specific}. Specifically, for each change of context in the input sequence, the specific changes to state variables are identified and translated to the corresponding MPs. Table \ref{tab:motion_primitives_general} is used to translate context changes in the general state variables while Table \ref{tab:motion_primitives_task_specific} is used to translate context changes in the task-specific state variables. 
If multiple states changed between labeled frames, then Grasp and Release MPs would have a higher priority than Touch and Untouch MPs (if they are performed on the same object by the same tool). Otherwise, all MPs were listed in the MP transcript so that separate MP transcripts for the left and right sides could be generated. 
Context labels are provided at 3 Hz and the context to MP translation assumes that states persist until the next context label in order to generate an MP label for each kinematic sample at 30 Hz.
Figure \ref{fig:translation} shows an example of a sequence of context translated into motion primitives.
This rule-based translation method assumes 
that changes in context can be completely described by the definitions in Tables \ref{tab:motion_primitives_general} and \ref{tab:motion_primitives_task_specific}. Alternatively, data-driven and learning from demonstration approaches can be used for more realistic and personalized modeling of the tasks and label translations.

\subsection{COMPASS Dataset}
\begin{figure}[b!]
    \centering
    \begin{minipage}[b]{.475\linewidth}
    \begin{subfigure}{\textwidth}
        \centering
        \includegraphics[trim = 3in 0in 3in 0.5in, clip, width=\textwidth]{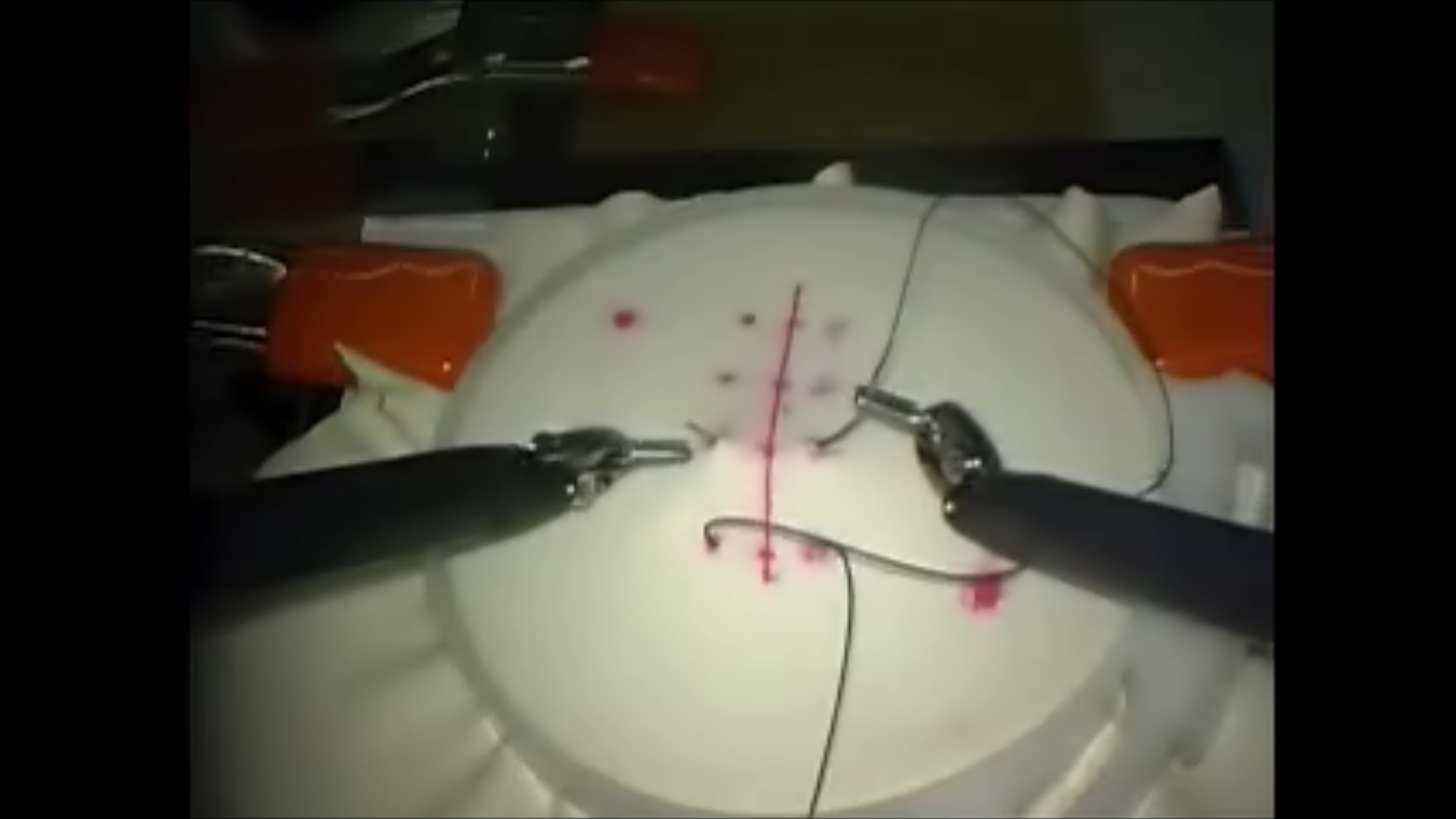}
        \caption{Suturing (S)}
        \label{fig:S_frame}
    \end{subfigure}
    \begin{subfigure}{\textwidth}
        \centering
        \includegraphics[trim = 3in 0.25in 3in 0.25in, clip, width=\textwidth]{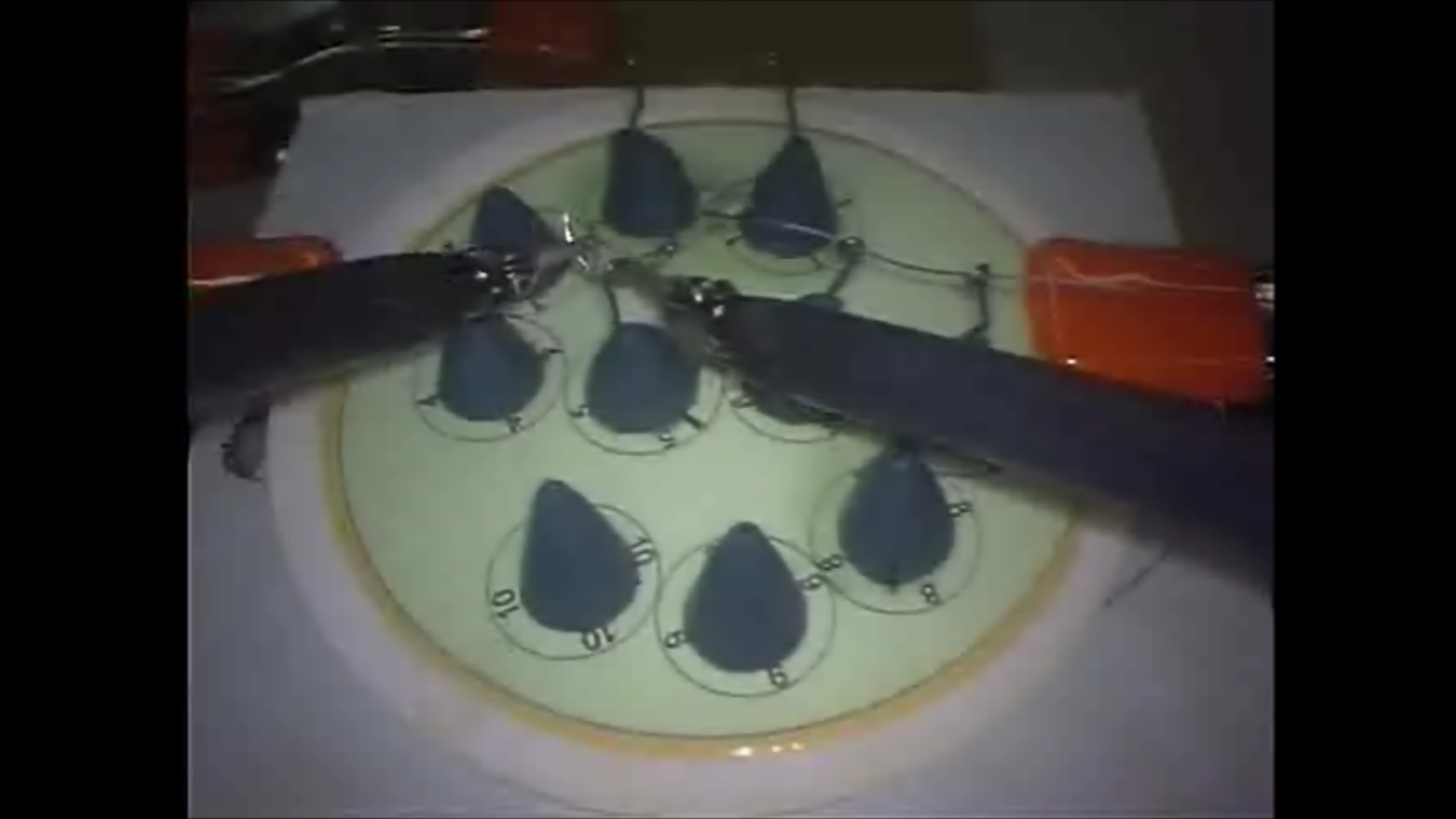}
        \caption{Needle Passing (NP)}
        \label{fig:NP_frame}
    \end{subfigure}
    \begin{subfigure}{\textwidth}
        \centering
        \includegraphics[trim = 3in 0.25in 3in 0.25in, clip, width=\textwidth]{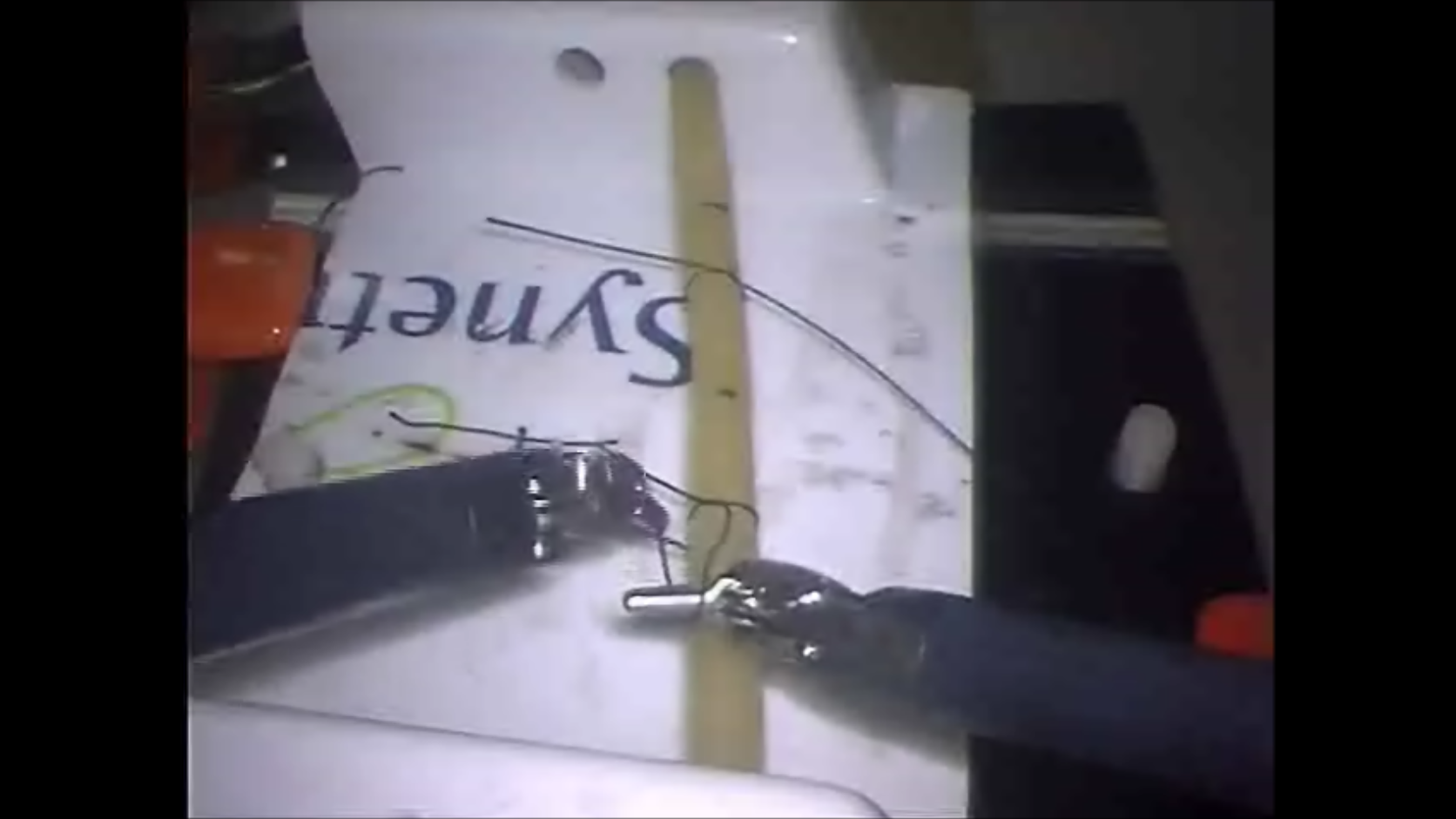}
        \caption{Knot Tying (KT)}
        \label{fig:KT_frame}
    \end{subfigure}
    \end{minipage}
    \begin{minipage}[b]{.475\linewidth}
    \begin{subfigure}{\textwidth}
        \centering
        \includegraphics[trim = 3in 0in 3in 0.5in, clip, width=\textwidth]{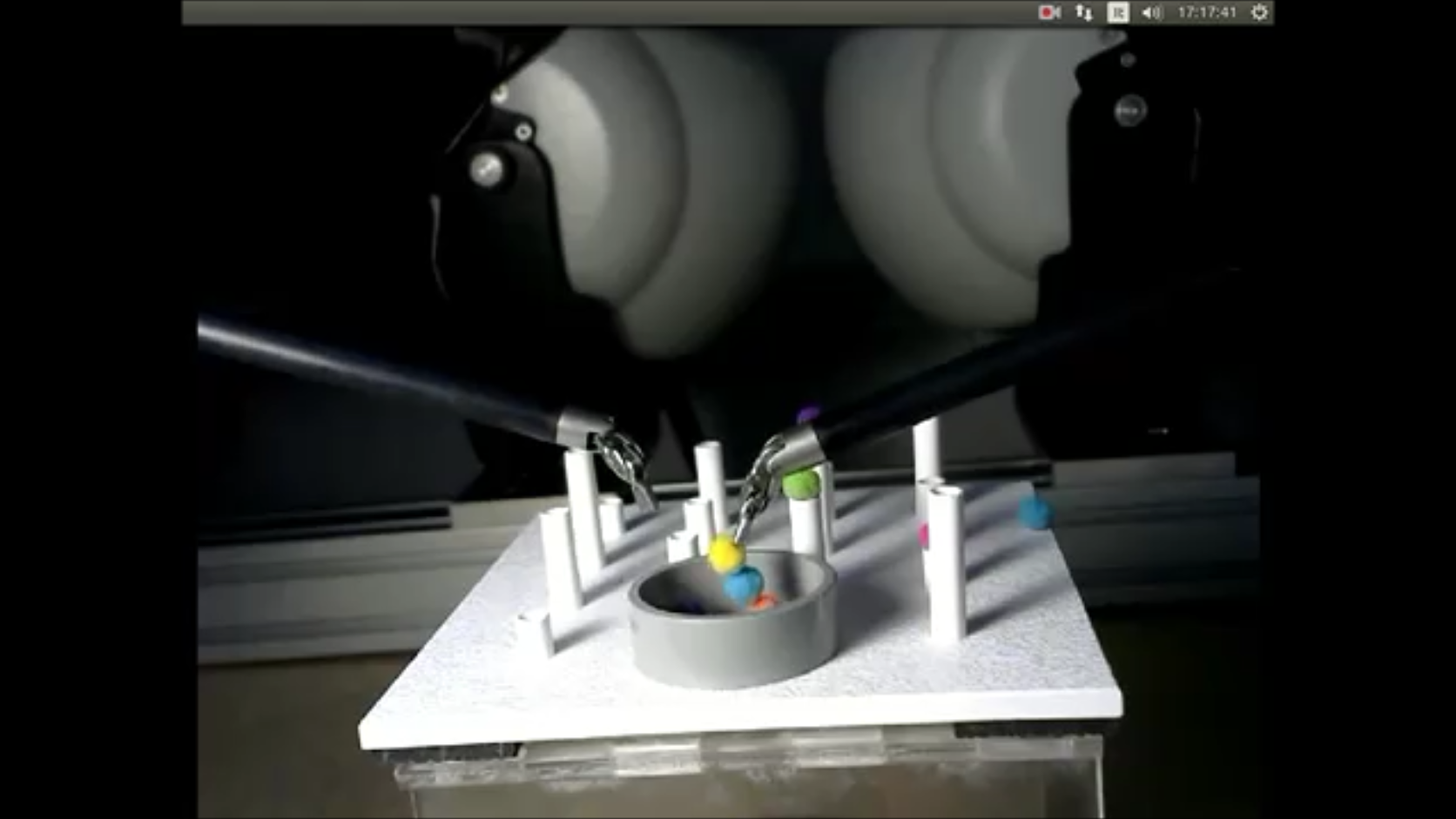}
        \caption{Pea on a Peg (PoaP)}
        \label{fig:PoaP_frame}
    \end{subfigure}
    \begin{subfigure}{\textwidth}
        \centering
        \includegraphics[trim = 3in 0in 3in 0.5in, clip, width=\textwidth]{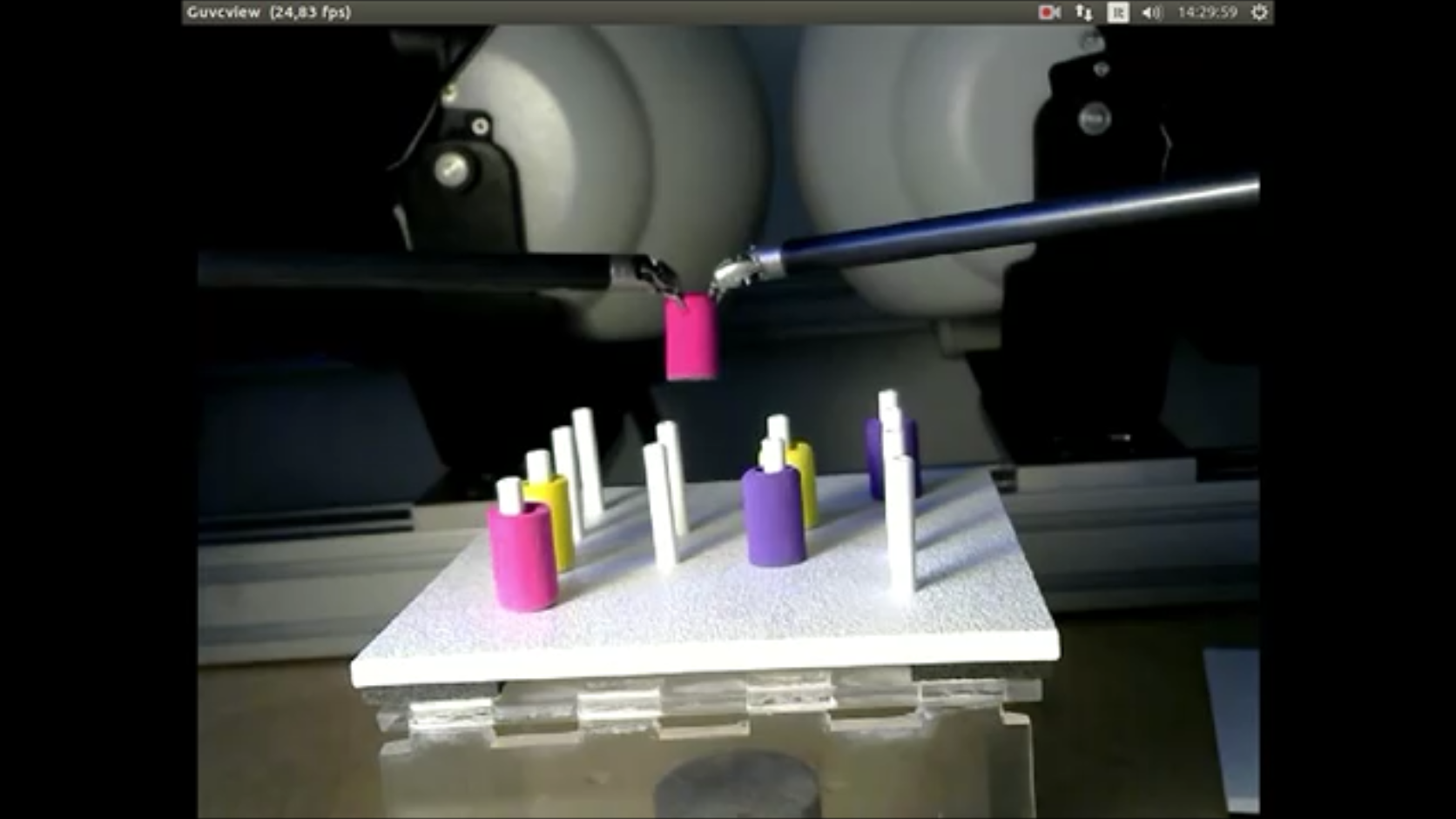}
        \caption{Post and Sleeve (PaS)}
        \label{fig:PaS_frame}
    \end{subfigure}
    \begin{subfigure}{\textwidth}
        \centering
        \includegraphics[trim = 3in 0.5in 3in 0in, clip, width=\textwidth]{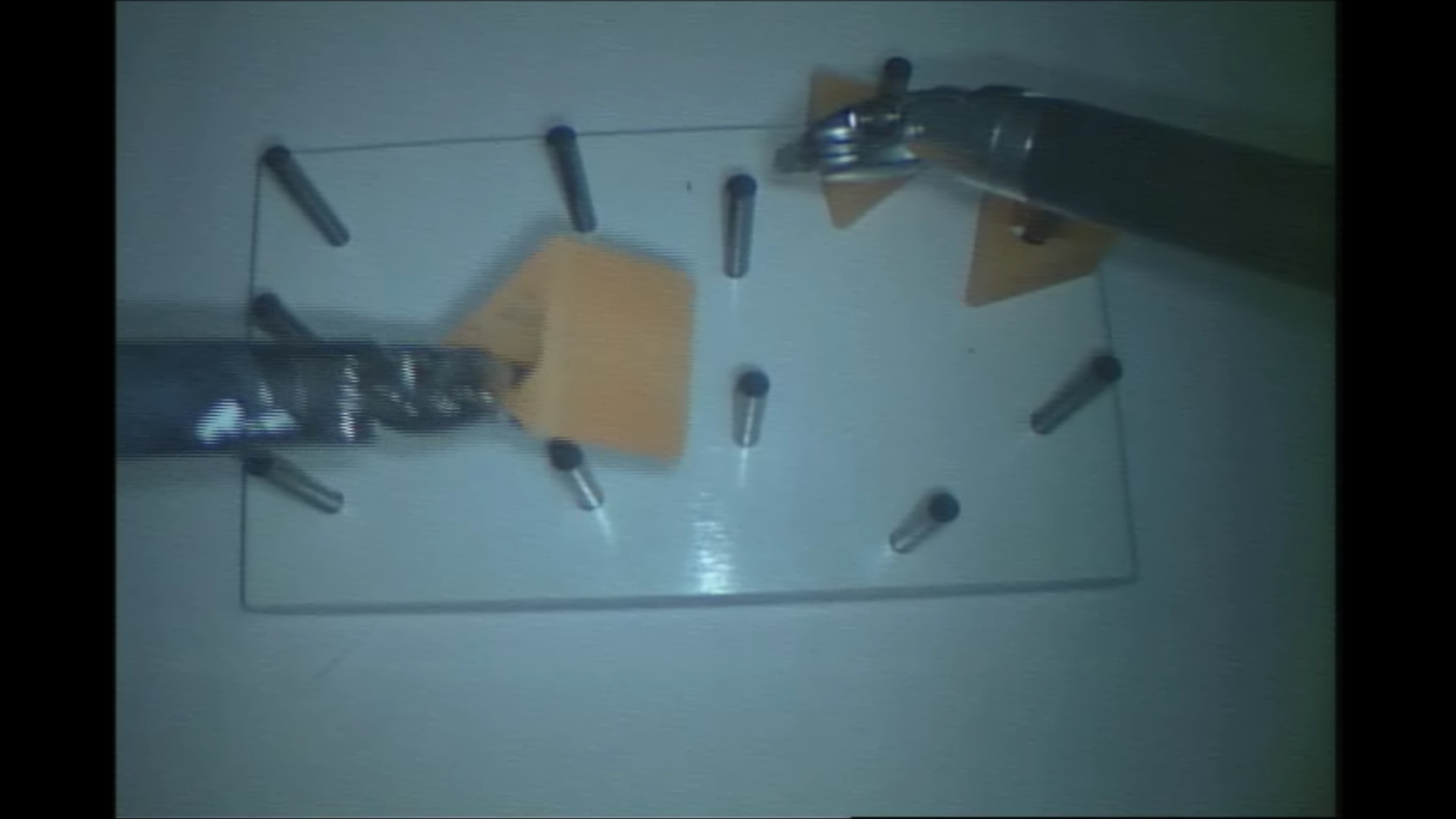}
        \caption{Peg Transfer (PT)}
        \label{fig:PT_frame}
    \end{subfigure}
    \end{minipage}
    
    \caption{Tasks included in the COMPASS dataset: Suturing (S), Needle Passing (NP), and Knot Tying (KT) from the JIGSAWS dataset \cite{gao2014jhu}; Pea on a Peg (PoaP) and Post and Sleeve (PaS) from the ROSMA dataset \cite{rivas2021surgical}; and Peg Transfer (PT) from the DESK dataset \cite{gonzalez2020desk}.}
    \label{fig:tasks}
    \vspace{-1.5em}
\end{figure}

We create the COMPASS dataset by aggregating data from 39 trials of Suturing (S), 28 trials of Needle Passing (NP), and 36 trials of Knot Tying (KT) performed by eight subjects from the JIGSAWS dataset; 47 trials of Peg Transfer (PT) performed by eight subjects from the DESK dataset; and 65 trials of Post and Sleeve (PaS), and 71 trials of Pea on a Peg (PoaP) performed by 12 subjects from the ROSMA dataset (see Figure \ref{fig:tasks}).

Videos are at 30 fps for the stereoscopic JIGSAWS and DESK tasks and 15 fps for the single camera for the ROSMA tasks. The kinematic data have been downsampled to 30 Hz and contain position, velocity, orientation (in quaternions), and gripper angle variables.
Since linear velocity data was not available for all tasks, it was derived from the position data using a rolling average over five samples. The ROSMA dataset did not contain gripper angle, so a separate round of manually labeling video data was performed to approximate the gripper angle as open or closed. To uniquely identify each individual file, a naming system that includes the task, subject number, and trial number was applied to facilitate matching kinematic, video, and label files. For example, the name Pea\_on\_a\_Peg\_S02\_T05 identifies the fifth trial of Pea on a Peg performed by subject two. 

To ensure reliable and high quality annotations, three full sets of context labels were obtained using our context labeling tool for all the trials. 
Two of the authors, with extensive experience with the datasets and the dry-lab robotic surgery tasks, each produced a full set of labels for all the trials. The third set of labels was crowd-sourced to engineering students. Consensus was then taken using majority voting for each state variable. 
We refer to this set as the ``Consensus" set. 

The COMPASS dataset includes these consensus context labels at 3 Hz and  the automatically generated motion primitive labels interpolated to 30 Hz for both arms of the robot so that every kinematic sample has an MP label. 
The original gesture labels from JIGSAWS and DESK datasets are synced and renamed under the new naming convention and included to promote comparisons between data and label sets.

\section{Evaluation}
In order to evaluate our context labeling and context to motion primitive translation methods, we obtain two more sets of labels, in addition to the ``Consensus" set. A group of expert surgeons labeled a set of six trials (one from each task) for context, referred to as the ``Surgeon" set, against which we evaluate the quality of the labels.
Three independent annotators also labeled a subset of trials in the JIGSAWS tasks for context, MPs, and gestures to assess and compare the different labeling methods and the context to motion primitive translation. We refer to these labels as the ``Multi-level" set.

\subsection{Context Labeling}
First, we assess the quality of context labels generated using our labeling tool by measuring the agreement among the annotators in ``Consensus" set as well as the agreement between the ``Surgeon" and ``Consensus" sets of context labels using Krippendorff's Alpha~\cite{park2012crowdsourcing}. Then, we compare context, MP, and gesture level labeling methods using labels in the ``Multi-level" set.

\textbf{Krippendorff's Alpha} is a commonly used statistical measure of inter-rater reliability and as shown in Equation \ref{equn:KA0} is calculated by considering the probability $ D_{e} $ that two labelers produced the same annotation due to chance rather than agreement on the data to label, and the observed disagreement $ D_{o} $ between each labeler's annotations:
\vspace{-0.5em}
\begin{equation}
    \alpha = 1 - \frac{D_o}{D_e}\\
\label{equn:KA0}
\end{equation}

\label{krippendorff}
The coefficient $\alpha$ is a value ranging from 1 to -1, with $\alpha > 0.8$ indicating near-perfect agreement, a value between 0.6 and 0.8 indicating substantial agreement, and smaller values indicating less agreement. $\alpha=0$ indicates no agreement other than by chance and negative values reflect more pronounced disagreement. 
Each of the labelers annotated a sequence of states encoded as numbers. The annotations do not have numerical significance and can best be described as categorical data, so the nominal distance or difference function is best suited to quantify the agreement between labelers annotating for context. The nominal distance or difference function in Equation \ref{equn:KA1} is used to calculate both $ D_{e} $ and $ D_{o} $ \cite{hughes2021krippendorffsalpha}, as given in Equation \ref{equn:KA2}, where $ n_l $ is the number of labelers  and $ n_u $ is the total number of frames which two or more labelers annotated.

\begin{equation}
  d_{nominal}(\text{label}_{1} \text{ label}_{2} )= 
  \begin{cases}
  0 & \text{if label$_{1}$ = label$_{2}$} \\ 
  1 & \text{if label$_{1}$ $\neq$ label$_{2}$} \\
  \end{cases}
\label{equn:KA1}
\end{equation}

\begin{equation}
\begin{aligned}
   D_{o} = \frac{1}{2 n_u n_l ( n_l - 1)} \sum_{l_{1}, l_{2} \in \text{all labels }}^{ } d_{nom}(l_{1}, l_{2} ) \\
   D_{e} =\frac{1}{2 n_u n_l (n_u n_l -1)} \sum_{l_{1}, l_{2} \in \text{all labels }}^{ } d_{nom}(l_{1}, l_{2} ) \\
\end{aligned}
\label{equn:KA2}
\end{equation}

\subsubsection{Consensus Context Labels}

The second column in Table \ref{tab:krippendorff_consensus} shows the agreement among crowd-sourced annotators, measured using the average Krippendorff's Alpha. In four of the tasks we see a near-perfect agreement ($\alpha$ above 0.8) and in two substantial agreement ($\alpha$ at least 0.6) among annotators. 
The average for all tasks was 0.84, weighted for the number of frames for each task, indicating substantial agreement in context labeling overall.
We observed that long segments of near-perfect agreement are punctuated by disagreements at the transitions between context. However, disagreement is limited to a few context states instead of the gesture label for a specific frame which results in much greater agreement between annotators when labeling for context than for gestures. Then, consensus obtained by separately majority voting for each state variable results in a high quality set of fine grained labels. 

The third column in Table \ref{tab:krippendorff_consensus} shows the Krippendorff's Alpha between Consensus and Surgeon context labels. 
We find that all tasks had an $\alpha$ of at least 0.8 and the average for all tasks  (weighted for the number of frames for each task) was 0.92, indicating near-perfect agreement between crowd-sourced context labels and those given by surgeons.

\subsubsection{Multi-level Labels}

Table \ref{tab:krippendorff_multi-level} shows the agreement among annotators for the ``Multi-level" labels. There is the least agreement when labeling using the descriptive gesture definitions, but labeling MPs directly is also difficult, likely due to their short durations. Annotating for context has the greatest agreement since labels are based on well-defined interactions among surgical tools and objects that are observed in video data.

Table \ref{tab:krippendorff_multi-level} also shows the agreement of the ``Multi-level" context and gesture labels with the ``Surgeon" context labels and JIGSAWS gesture labels, respectively. There is much higher agreement when labeling for context than for gestures and the existing JIGSAWS labels are difficult to reproduce. This might be because these labels were generated more subjectively by only one annotator through watching the videos and in consultation with a surgeon~\cite{JIGSAWS}.
We also again see that crowd-sourcing context labels results in high quality annotations which are comparable to those given by expert surgeons.

\begin{table}[t!]
\centering
\caption{Krippendorff's Alpha among annotators and between Consensus and Surgeon context labels.}
\label{tab:krippendorff_consensus}
\begin{tabular}{P{2.5cm} P{2cm} P{3cm}}
\toprule
Task & Among annotators & Between Consensus and Surgeon \\ \midrule
Suturing & 0.69 & 0.86 \\ 
Needle Passing & 0.85 & 0.90 \\ 
Knot Tying & 0.79 & 0.94 \\ 
Peg Transfer & 0.90 & 0.94 \\ 
Pea on a Peg & 0.83 & 0.93 \\ 
Post and Sleeve & 0.89 & 0.97 \\ 
\bottomrule
\end{tabular}
\vspace{-1em}
\end{table}

\begin{table}[t!]
\centering
\caption{Krippendorff's Alpha among annotators for context, MP, and gesture Multi-level labels.}
\label{tab:krippendorff_multi-level}
\begin{tabular}{P{0.5cm} P{1cm} P{1cm} P{1.2cm} | P{2.75cm} P{2.75cm}}
\toprule
\multirow{2}{0.5cm}{Task} & \multicolumn{3}{P{3.2cm}|}{Multi-Level} & \multirow{2}{2.75cm}{Multi-Level vs. Surgeon Context} & \multirow{2}{2.75cm}{Multi-Level vs. JIGSAWS Gestures}  \\
 & Context & MPs & Gestures & &  \\ \midrule
S & 0.72 & 0.33 & 0.24  & 0.86 & 0.34 \\ 
NP & 0.91 & 0.41 & 0.08 & 0.90 & 0.04 \\ 
KT & 0.89 & 0.26 & 0.20 & 0.89 & 0.06 \\ 
\bottomrule
\end{tabular}
\end{table}

\subsection{Context to Motion Primitive Translation} 
To assess the performance of the context to MP translation, we translate the context labels in the ``Multi-level" annotations set and compare the resulting translated MP transcripts to the ground truth MP labels for each annotator. We calculate the accuracy and edit score for each task as described below.

\textbf{Accuracy:}
Given the lists of predicted and ground truth labels, the accuracy is the ratio of correctly classified samples divided by the total number of samples in a trial.

\textbf{Edit Score:}
We report the edit score as defined in \cite{lea2016temporal} which uses the normalized Levenshtein edit distance, $ edit(G, P) $, by calculating the number of insertions, deletions, and replacements needed to transform the sequence of predicted labels $ P $  to match the ground truth sequence of labels $ G $. The edit score is normalized by the maximum length of the predicted and ground truth sequences and is thus computed using Equation \ref{equn:edit} where 100 is the best and 0 is the worst.

\begin{equation}
    \text{Edit Score} = (1-\frac{edit(G, P)}{max(len(G), len(P))}) \times 100
    \label{equn:edit}
\end{equation}

\subsubsection{Quality of Multi-level Labels}
The input context labels and the ground truth MP labels show variability by the type of task (S, NP, KT) and by the skill of the annotator, both of which can affect the resulting translated MP labels and their evaluation. So we first used the agreement scores with the ``Surgeon" context labels set and JIGSAWS gesture labels as a measure to assess the quality of each annotator and found appreciable differences in the accuracy of each annotator. The accuracies for annotators 1, 2, and 3 were, respectively, 0.58, 0.69, and 0.71 when labeling for context (compared to surgeons), and 0.27, 0.35, and 0.65 when labeling for gestures (compared to JIGSAWS). 
Thus, annotator 3 was the most reliable annotator overall, but annotator 2 was almost as reliable for context labels. We also found that the annotators had similar agreement with surgeons for NP which is consistent with the results in Table \ref{tab:krippendorff_consensus}. 

\begin{table}[b!]
\centering
\vspace{-1em}
\caption{Accuracy and edit score between consensus ground truth and translated motion primitives for Multi-level labels.}
\label{tab:translation_metrics}
\begin{tabular}{P{1.4cm} P{0.70cm} P{0.70cm} P{0.70cm} P{0.70cm} P{0.70cm} P{0.70cm}}
\toprule
 &  \multicolumn{2}{c}{\textbf{Annotator 1}} & \multicolumn{2}{c}{\textbf{Annotator 2}} & \multicolumn{2}{c}{\textbf{Annotator 3}} \\ 
Task &  \multicolumn{2}{c}{\text{(Acc/Edit Score)}} & \multicolumn{2}{c}{\text{(Acc/Edit Score)}} & \multicolumn{2}{c}{\text{(Acc/Edit Score)}} \\ \midrule
  
S       & 0.27 & 33.9 & 0.18 & 26.9 & 0.23 & 30.1 \\
NP & 0.64 & 67.4 & 0.27 & 45.1 & 0.45 & 46.4 \\
KT     & 0.50 & 53.8 & 0.31 & 56.1 & 0.53 & 56.8 \\  
\bottomrule
\end{tabular}
\end{table}

\subsubsection{Translation Accuracy}
As shown in Table \ref{tab:translation_metrics}, the context to motion primitive translation accuracy was higher for annotators 1 and 3 compared to annotator 2. Also the task breakdown reveals inter-rater variability across tasks, with highest edit score for annotator 1 on S and NP tasks and for annotator 3 on KT.

However, the ground truth MP labels used in this evaluation had very low agreement among annotators compared to context labels and assessing their reliability is beyond the scope of this paper. Future work will collaborate with surgeons for 
generating high quality annotations using multi-level labeling methods so we can better evaluate these different methods and improve the automated translation between them.

\section{Related Work}
\label{sect:related_work}
The ``Language of Surgery" project \cite{lin2010structure} models surgical procedures as a language and uses grammar to dictate how gestures are combined to perform tasks. A hierarchical framework has been proposed to model surgical procedures \cite{neumuth2011modeling}. Available datasets primarily focus on the task, gesture, and action levels as summarized in Table \ref{tab:related_work_gestures} and Table \ref{tab:related_work_actions}. Within this hierarchical framework, as shown in Section \ref{sec:Framework}, tasks consist of a sequence of gestures, and actions are defined below the gesture level. 

\textbf{Surgical gestures:} Gestures are defined as ``the smallest  surgical motion gesture that encapsulates a specific intent, (e.g., insert needle through tissue)" with semantic meaning. The JIGSAWS dataset \cite{gao2014jhu}, provides gesture level labels for the Suturing, Needle Passing, and Knot Tying tasks in a dry lab experiment setting. Recent works have introduced new datasets as shown in Table \ref{tab:related_work_gestures}, but differing gesture definitions limit comparisons between them as well as their generalizability to other tasks. Specifically, \cite{dipietro2019segmenting}, \cite{menegozzo2019surgical}, and \cite{goldbraikh2022using} all performed gesture recognition based on kinematic data, but used different datasets and gesture definitions making comparisons difficult. \cite{qin2020davincinet} fused kinematic, video, and event data to recognize and predict gestures. But previous works have not combined data from multiple sets since the gesture labels were incompatible.

\textbf{Action triplets:} Action triplets, $<$surgical tool /instrument, action verb, target anatomy$>$, are used to describe  tool-tissue interactions (TTI) in surgical process modeling \cite{Nwoye2020recginter}. One of the early works formalizes Laparoscopic Adreanectomies, Cholecystectomies and Pancreatic Resections \cite{Kati2014model} with
surgical activities in the form of action triplets for surgical phase inference. \cite{Xu2021report} annotates two robotic surgery
datasets of MICCAI robotic scene segmentation and Transoral
Robotic Surgery (TORS) in the form of action triplets to generate surgical reports. In the SARAS Endoscopic Surgeon Action Detection
(ESAD) challenge \cite{singh2021saras}, instead of action triplets, actions are described by both the verb and the anatomy. In the CholecTriplet2021 benchmark challenge for surgical
action triplet recognition \cite{nwoye2022cholectriplet2021}, the challenge dataset, CholecT50, consists
of 50 video recordings of laparoscopic cholecystectomy labeled for 100 action triplet classes composed from 6 instruments, 10
verbs, and 15 targets. Despite being more descriptive of the surgical scene, the number of action triplets in the form of verbs, instruments, and targets can grow exponentially compared to a more limited number of gestures. 

\textbf{Surgical Actions:} Surgical actions are generally referred to as the level of the surgical hierarchy below gestures. Motions, motion primitives, and the action verb in action triplets are surgical actions. In surgical process modeling, \cite{neumuth2011modeling} and \cite{lalys2014surgical} define motions as an activity performed by only one hand and without semantic meaning. Many other works 
define actions as atomic units as listed in Table \ref{tab:related_work_actions}.  

The gesture and action label datasets are mostly proposed for surgical workflow segmentation \cite{twinanda2016endonet, valderrama2022towards, ward2020training, wagner2021comparative, ban2021surgical}, and gesture  \cite{dipietro2019segmenting,menegozzo2019surgical,goldbraikh2022using,qin2020davincinet}, action, or action triplet recognition \cite{wagner2021comparative, nwoye2022rendezvous,li2022sirnet}. However, different datasets have varying definitions of gestures and actions. This makes
combining data from multiple sets challenging. 
Besides varying definitions, prior datasets on surgical actions mainly contain video data, and none of the prior datasets look into the process of labeling and labeling agreement. Prior datasets also do not differentiate actions performed by the left and right hands, which are important for detailed skill assessment and analysis of bimanual coordination. In our framework, we formally define motion primitives for the left and right hands. Our motion primitives are sets of action definitions that are generalizable across different datasets. We also look into how motion primitives relate to surgical context and task progress. Our proposed dataset contains both kinematic and video data along with context and motion primitives labels for a total of six dry-lab tasks. Our dataset will facilitate the development of recognition, skill assessment, and error detection models using both vision and kinematic data.

\noindent
\begin{table}[h!]
\begin{center}
\begin{minipage}{\textwidth} 
\caption{Datasets and definitions for gesture recognition}
\label{tab:related_work_gestures}
\resizebox{\textwidth}{!}{
\begin{tabular}{p{2.5cm} p{2cm} p{2.25cm} p{9cm}}
\toprule
Paper & Dataset & Tasks & Gestures \\
\midrule

Gao 2014 \cite{gao2014jhu} \cite{ahmidi2017dataset} & \begin{tabular}[t]{@{} p{2cm}} JIGSAWS \\ \begin{itemize}[leftmargin=0.4cm, before=\vspace{-0.75em}] \item Video  \item Kinematics  \end{itemize} \end{tabular} & \begin{tabular}[t]{@{} p{2.25cm}} Suturing \\ Needle Passing \\ Knot Tying \end{tabular} & \begin{tabular}[t]{@{} p{9cm}} G1 - Reaching for needle with right hand \\ G2 - Positioning needle \\ G3 - Pushing needle through tissue \\ G4 - Transferring needle from left to right \\ G5 - Moving to center with needle in grip \\ G6 - Pulling suture with left hand \\ G7 - Pulling suture with right hand \\ G8 - Orienting needle \\ G9 - Using right hand to help tighten suture \\ G10 - Loosening more suture \\ G11 - Dropping suture at end and moving to end points \\ G12 - Reaching for needle with left hand \\ G13 - Making C loop around right hand \\ G14 - Reaching for suture with right hand \\ G15 - Pulling suture with both hands \end{tabular} \\
\hline

DiPietro 2019 \cite{dipietro2019segmenting} &  \begin{tabular}[t]{@{} p{2cm}} MISTIC-SL \\ \begin{itemize}[leftmargin=0.4cm, before=\vspace{-0.75em}] \item Video  \item Kinematics  \end{itemize} \end{tabular} & \begin{tabular}[t]{@{} p{2.25cm}} Suturing \\ Needle Passing \\ Knot Tying \end{tabular} & \begin{tabular}[t]{@{} p{9cm}} G1-G12 \& G14 \\ G13 - Grab suture using 2nd needle driver \\ G15 - Rotate suture twice using 1st needle driver around 2nd needle driver \\ G16 - Grab suture tail using 2nd needle driver in knot tying \\ G17 - Pull suture tail using 2nd needle driver through knot \\ G18 - Pull ends of suture taut \\ G19 - Rotate suture once using 2nd needle driver around 1st needle driver \\ G20 - Grab suture tail using 1st needle driver in knot tying \\ G21 - Pull suture tail using 1st needle driver through knot \\ G22 - Grab suture using 1st needle driver \end{tabular} \\
\hline

Gonzalez 2020 \cite{gonzalez2020desk} & \begin{tabular}[t]{@{} p{2cm}} DESK \\ \begin{itemize}[leftmargin=0.4cm, before=\vspace{-0.75em}] \item Video  \item Kinematics  \end{itemize} \end{tabular} & \begin{tabular}[t]{@{} p{2.25cm}} Peg Transfer \end{tabular} & \begin{tabular}[t]{@{} p{4.5cm} p{4.5cm}} S1 - Approach peg & S5 - Transfer peg - Exchange \\ S2 - Align \& grasp & S6 - Approach pole \\ S3 - Lift peg & S7 - Align \& place \\ S4 - Transfer peg - Get together &  \\ \end{tabular} \\
\hline

Menegozzo 2019 \cite{menegozzo2019surgical} & \begin{tabular}[t]{@{} p{2cm}} V-RASTED \\ \begin{itemize}[leftmargin=0.4cm, before=\vspace{-0.75em}, after=\vspace{-1em}] \item Video  \item Kinematics  \end{itemize} \end{tabular} & \begin{tabular}[t]{@{} p{2.25cm}} Pick and Place\end{tabular} & \begin{tabular}[t]{@{} p{4.5cm} p{4.5cm}} 1 – Collecting ring & 4 – Failing 1 \\ 2 – Passing ring R to L & 5 – Failing 2\\ 3 – Posing ring on pole & 6 – Failing 3 \end{tabular} \\ 
\hline

Goldbraikh 2022 \cite{goldbraikh2022using} & \begin{tabular}[t]{@{} p{2cm}} own \\ \begin{itemize}[leftmargin=0.4cm, before=\vspace{-0.75em}, after=\vspace{-1em}] \item Video \item Kinematics  \end{itemize} \end{tabular} & \begin{tabular}[t]{@{} p{2.25cm}} Suturing \\ (not robotic) \end{tabular} & \begin{tabular}[t]{@{} p{4.5cm} p{4.5cm}} No gesture & Instrument tie \\ Needle passing & Lay the knot \\ Pull the suture & Cut the suture\end{tabular} \\ 
\hline

Qin 2020 \cite{qin2020temporal} & \begin{tabular}[t]{@{} p{2cm}} RIOUS \\ \begin{itemize}[leftmargin=0.4cm, before=\vspace{-0.75em}, after=\vspace{-1em}] \item Kinematics \item Video \item Events \end{itemize} \end{tabular} & \begin{tabular}[t]{@{} p{2.25cm}} Ultrasonic probing \end{tabular} & \begin{tabular}[t]{@{} p{4.5cm} p{4.5cm}} S1 Probe released, out of view & S5 Lifting probe up \\ S2 Probe released, in view & S6 Carrying probe to tissue surface \\ S3 Reaching for probe & S7 Sweeping \\ S4 Grasping probe & S8 Releasing probe \end{tabular} \\

\botrule
\end{tabular}
}
\end{minipage}
\end{center}
\end{table}

\noindent
\begin{table}[h!]
\begin{center}
\begin{minipage}{\textwidth} 
\caption{Datasets and models for surgical actions}
\label{tab:related_work_actions}
\resizebox{\textwidth}{!}{
\begin{tabular}{p{2.5cm} p{2cm} p{2.25cm} p{9cm}}
\toprule
Paper & Dataset & Tasks & Actions \\
\midrule
Nwoye 2022 \cite{nwoye2022rendezvous} & \begin{tabular}[t]{@{} p{2cm}} CholecT50 \\ \begin{itemize}[leftmargin=0.4cm, before=\vspace{-0.75em}] \item Video  \end{itemize} \end{tabular} & \begin{tabular}[t]{@{} p{2.25cm}}  Laparoscopic cholecystectomy \end{tabular} & \begin{tabular}[t]{@{} p{3cm} p{3cm} p{3cm}}  Aspirate & Dissect & Pack \\ Clip & Grasp & Retract \\ Coagulate & Irrigate &  \\ Cut & Null & \\\end{tabular} \\ 
\hline

Li 2022 \cite{li2022sirnet} &\begin{tabular}[t]{@{} p{2cm}} EndoVis2018 \\ \begin{itemize}[leftmargin=0.4cm, before=\vspace{-0.75em}] \item Video \end{itemize} \end{tabular} & \begin{tabular}[t]{@{} p{2.25cm}} Nephrectomy \end{tabular} & \begin{tabular}[t]{@{} p{3cm} p{3cm} p{3cm}} Cauterization & Looping & Clipping \\ Suction & Idle & Retraction \\ Staple & Tool manipulation & \\ Ultrasound sensing & Suturing & \end{tabular} \\ 
\hline

Meli 2021 \cite{meli2021unsupervised} & \begin{tabular}[t]{@{} p{2cm}} own \\ \begin{itemize}[leftmargin=0.4cm, before=\vspace{-0.75em}] \item Video \item Kinematics  \end{itemize} \end{tabular} & \begin{tabular}[t]{@{} p{2.25cm}} Ring Transfer \end{tabular} & \begin{tabular}[t]{@{} p{3cm}} Move \\ Grasp \\ Release \\ Extract \end{tabular} \\ 
\hline

Forestier 2012 \cite{forestier2012classification} & \begin{tabular}[t]{@{} p{2cm}} own \\ \begin{itemize}[leftmargin=0.4cm, before=\vspace{-0.75em}] \item Video \end{itemize} \end{tabular} & \begin{tabular}[t]{@{} p{2.25cm}} Lumbar disk \\ herniation \end{tabular} & \begin{tabular}[t]{@{} p{3cm} p{3cm}} Right: & Left:\\ Sew & Hold \\ Install & Install \\ Hold & Remove\\ Remove & \\ Coagulate & \\ Swab & \\Irrigate & \end{tabular} \\ 
\hline

Wagner 2021 \cite{wagner2021comparative} & \begin{tabular}[t]{@{} p{2cm}} EndoVis 2019 \\ \begin{itemize}[leftmargin=0.4cm, before=\vspace{-0.75em}] \item Video  \end{itemize} \end{tabular} & \begin{tabular}[t]{@{} p{2.25cm}}  Laparoscopic cholecystectomy \end{tabular} & \begin{tabular}[t]{@{} p{3cm}} Grasp \\  Hold \\ Cut \\ Clip \end{tabular} \\ 
\hline

De Rossi 2021 \cite{de2021first} & \begin{tabular}[t]{@{} p{2cm}} own \\ \begin{itemize}[leftmargin=0.4cm, before=\vspace{-0.75em}] \item Video  \end{itemize} \end{tabular} & \begin{tabular}[t]{@{} p{2.25cm}} Pick and place \\ (semi-autonomous and cooperative) \end{tabular} & \begin{tabular}[t]{@{} p{9cm}}A01 – MS moves to ring\\ A02 – MS picks ring\\ A03 – MS moves ring to exchange area\\ A04 – AS moves to ring\\ A05 – AS grasps ring and MS leaves ring\\ A06 – AS moves ring to delivery area\\ A07 – AS drops ring on target\\ A08 – AS moves to starting position \end{tabular} \\
\hline


Valderrama 2022 \cite{valderrama2022towards} & \begin{tabular}[t]{@{} p{2cm}} PSI-AVA \\ \begin{itemize}[leftmargin=0.4cm, before=\vspace{-0.75em}] \item Video  \end{itemize} \end{tabular} & \begin{tabular}[t]{@{} p{2.25cm}}  Radical prostatectomy \end{tabular} & \begin{tabular}[t]{@{} p{3cm} p{3cm} p{3cm}} Cauterize & Open & Still \\ Close & Open Something & Suction \\ Close Something & Pull & Travel \\ Cut & Push & Wash \\ Grasp & Release &  \\ Hold & Staple & \end{tabular} \\ 
\hline 

Ma 2021 \cite{ma2021novel} & \begin{tabular}[t]{@{} p{2cm}} own \\  \begin{itemize}[leftmargin=0.4cm, before=\vspace{-0.75em}] \item Video \end{itemize} \end{tabular} & \begin{tabular}[t]{@{} p{2.25cm}} Renal hilum \\ dissection \\ (Partial nephrectomy) \end{tabular} & \begin{tabular}[t]{@{} p{3cm} p{3cm} p{3cm}} Single blunt dissection: & Single sharp dissection: & Combination: \\ Spread & Cold cut & Pedicalize \\ Peel/push & Hot cut & 2-hand spread \\ Hook & Burn dissect & Coagulate then cut \end{tabular} \\ 
\hline 

Huaulm{\`e} 2021 \cite{huaulme2021micro} & \begin{tabular}[t]{@{} p{2cm}} MISAW \\  \begin{itemize}[leftmargin=0.4cm, before=\vspace{-0.75em}] \item Kinematic \item Video \end{itemize} \end{tabular} & \begin{tabular}[t]{@{} p{2.25cm}} Suturing \\ Knot Tying \end{tabular} & \begin{tabular}[t]{@{} p{3cm} p{3cm} p{3cm}} Catch  & Loosen completely & Pass through \\ Give slack & Loosen partially & Position \\ Hold & Make a loop & Pull \\ Insert & & \\ \end{tabular} \\ 

\botrule
\end{tabular}
}
\end{minipage}
\end{center}
\end{table}

\section{Discussion and Conclusion}

In summary, we present a framework for modeling surgical tasks as finite state machines where motion primitives cause changes in surgical context. We apply our framework to three publicly available datasets to create an aggregate dataset of kinematic and video data along with context and motion primitive labels. Our method for labeling context achieves substantial to near-perfect agreement between annotators and expert surgeons. 
Using motion primitives, we aggregate data from different datasets, tasks, and subjects and nearly triple the amount of data with consistent label definitions. 

Future work includes extending the motion primitive framework to tasks from real surgical procedures which would be accomplished by defining task-specific state variables to augment the context labels and their associated motion primitives. 
Our standardized set of context and motion primitive labels enables the generalized modeling and comparison of surgical activities between datasets and tasks. This supports the development of models for surgical activity recognition, skill analysis, error detection, and surgical automation. 

\backmatter





\bmhead{Acknowledgments}

This work was supported in part by the National Science Foundation grants DGE-1842490, DGE-1829004, and CNS-2146295 and by the Engineering-in-Medicine center at the University of Virginia. We thank the volunteer labelers and Dr. Schenkman, Dr. Cantrell, and Dr. Chen for their medical feedback.

\section*{Declarations}
\subsection*{Competing Interests}
The authors declare that they have no conflict of interest.
\subsection*{Ethics Approval}
This article does not contain any studies involving human participants performed by any of the authors.
\subsection*{Informed Consent}
This article does not contain any studies involving human participants performed by any of the authors.

\bibliographystyle{plain}
\bibliography{main.bib}

\begin{thebibliography}{10}

\bibitem{ahmidi2017dataset}
Narges Ahmidi, Lingling Tao, Shahin Sefati, Yixin Gao, Colin Lea,
  Benjamin~Bejar Haro, Luca Zappella, Sanjeev Khudanpur, Ren{\'e} Vidal, and
  Gregory~D Hager.
\newblock A dataset and benchmarks for segmentation and recognition of gestures
  in robotic surgery.
\newblock {\em IEEE Transactions on Biomedical Engineering}, 64(9):2025--2041,
  2017.

\bibitem{allan20202018}
Max Allan, Satoshi Kondo, Sebastian Bodenstedt, Stefan Leger, Rahim
  Kadkhodamohammadi, Imanol Luengo, Felix Fuentes, Evangello Flouty, Ahmed
  Mohammed, Marius Pedersen, Avinash Kori, Varghese Alex, Ganapathy
  Krishnamurthi, David Rauber, Robert Mendel, Christoph Palm, Sophia Bano,
  Guinther Saibro, Chi-Sheng Shih, Hsun-An Chiang, Juntang Zhuang, Junlin Yang,
  Vladimir Iglovikov, Anton Dobrenkii, Madhu Reddiboina, Anubhav Reddy,
  Xingtong Liu, Cong Gao, Mathias Unberath, Myeonghyeon Kim, Chanho Kim,
  Chaewon Kim, Hyejin Kim, Gyeongmin Lee, Ihsan Ullah, Miguel Luna, Sang~Hyun
  Park, Mahdi Azizian, Danail Stoyanov, Lena Maier-Hein, and Stefanie Speidel.
\newblock 2018 robotic scene segmentation challenge.
\newblock {\em arXiv preprint arXiv:2001.11190}, 2020.

\bibitem{ban2021surgical}
Yutong Ban, Guy Rosman, Thomas Ward, Daniel Hashimoto, Taisei Kondo, Hidekazu
  Iwaki, Ozanan Meireles, and Daniela Rus.
\newblock Surgical prediction gan for events anticipation.
\newblock {\em arXiv preprint arXiv:2105.04642}, 2021.

\bibitem{boehm2021online}
Jacob~R Boehm, Nicholas~P Fey, and Ann~Majewicz Fey.
\newblock Online recognition of bimanual coordination provides important
  context for movement data in bimanual teleoperated robots.
\newblock In {\em 2021 IEEE/RSJ International Conference on Intelligent Robots
  and Systems (IROS)}, pages 6248--6255. IEEE, 2021.

\bibitem{bowyer2013active}
Stuart~A Bowyer, Brian~L Davies, and Ferdinando~Rodriguez y~Baena.
\newblock Active constraints/virtual fixtures: A survey.
\newblock {\em IEEE Transactions on Robotics}, 30(1):138--157, 2013.

\bibitem{JIGSAWS}
{Computational Interaction and Robotics Laboratory}.
\newblock {JHU-ISI Gesture and Skill Assessment Working Set (JIGSAWS)}.
\newblock
  \url{https://cirl.lcsr.jhu.edu/research/hmm/datasets/jigsaws_release/}.

\bibitem{de2021first}
Giacomo De~Rossi, Marco Minelli, Serena Roin, Fabio Falezza, Alessio Sozzi,
  Federica Ferraguti, Francesco Setti, Marcello Bonf{\`e}, Cristian Secchi, and
  Riccardo Muradore.
\newblock A first evaluation of a multi-modal learning system to control
  surgical assistant robots via action segmentation.
\newblock {\em IEEE Transactions on Medical Robotics and Bionics}, 2021.

\bibitem{dipietro2019segmenting}
Robert DiPietro, Narges Ahmidi, Anand Malpani, Madeleine Waldram, Gyusung~I
  Lee, Mija~R Lee, S~Swaroop Vedula, and Gregory~D Hager.
\newblock Segmenting and classifying activities in robot-assisted surgery with
  recurrent neural networks.
\newblock {\em International journal of computer assisted radiology and
  surgery}, 14(11):2005--2020, 2019.

\bibitem{falezza2021modeling}
Fabio Falezza, Nicola Piccinelli, Giacomo De~Rossi, Andrea Roberti, Gernot
  Kronreif, Francesco Setti, Paolo Fiorini, and Riccardo Muradore.
\newblock Modeling of surgical procedures using statecharts for semi-autonomous
  robotic surgery.
\newblock {\em IEEE Transactions on Medical Robotics and Bionics},
  3(4):888--899, 2021.

\bibitem{forestier2012classification}
Germain Forestier, Florent Lalys, Laurent Riffaud, Brivael Trelhu, and Pierre
  Jannin.
\newblock Classification of surgical processes using dynamic time warping.
\newblock {\em Journal of biomedical informatics}, 45(2):255--264, 2012.

\bibitem{gao2014jhu}
Yixin Gao, S~Swaroop Vedula, Carol~E Reiley, Narges Ahmidi, Balakrishnan
  Varadarajan, Henry~C Lin, Lingling Tao, Luca Zappella, Benjam{\i}n B{\'e}jar,
  David~D Yuh, Chi Chiung~Grace Chen, Ren{\'e} Vidal, Sanjeev Khudanpur, and
  Gregory~D. Hager.
\newblock Jhu-isi gesture and skill assessment working set (jigsaws): A
  surgical activity dataset for human motion modeling.
\newblock In {\em MICCAI Workshop: M2CAI}, volume~3, page~3, 2014.

\bibitem{ginesi2020autonomous}
Michele Ginesi, Daniele Meli, Andrea Roberti, Nicola Sansonetto, and Paolo
  Fiorini.
\newblock Autonomous task planning and situation awareness in robotic surgery.
\newblock In {\em 2020 IEEE/RSJ International Conference on Intelligent Robots
  and Systems (IROS)}, pages 3144--3150. IEEE, 2020.

\bibitem{ginesi2019dmp++}
Michele Ginesi, Nicola Sansonetto, and Paolo Fiorini.
\newblock Dmp++: Overcoming some drawbacks of dynamic movement primitives.
\newblock {\em arXiv preprint arXiv:1908.10608}, 2019.

\bibitem{ginesi2021overcoming}
Michele Ginesi, Nicola Sansonetto, and Paolo Fiorini.
\newblock Overcoming some drawbacks of dynamic movement primitives.
\newblock {\em Robotics and Autonomous Systems}, 144:103844, 2021.

\bibitem{goldbraikh2022using}
Adam Goldbraikh, Tomer Volk, Carla~M Pugh, and Shlomi Laufer.
\newblock Using open surgery simulation kinematic data for tool and gesture
  recognition.
\newblock {\em International Journal of Computer Assisted Radiology and
  Surgery}, pages 1--15, 2022.

\bibitem{gonzalez2020desk}
Glebys~T Gonzalez, Upinder Kaur, Masudur Rahma, Vishnunandan Venkatesh, Natalia
  Sanchez, Gregory Hager, Yexiang Xue, Richard Voyles, and Juan Wachs.
\newblock From the desk (dexterous surgical skill) to the battlefield--a
  robotics exploratory study.
\newblock {\em arXiv preprint arXiv:2011.15100}, 2020.

\bibitem{huaulme2021micro}
Arnaud Huaulm{\'e}, Duygu Sarikaya, K{\'e}vin Le~Mut, Fabien Despinoy, Yonghao
  Long, Qi~Dou, Chin-Boon Chng, Wenjun Lin, Satoshi Kondo, Laura
  Bravo-S{\'a}nchez, Pablo Arbel{\'a}ez, Wolfgang Reiter, Manoru Mitsuishi,
  Kanako Harada, and Pierre Jannin.
\newblock Micro-surgical anastomose workflow recognition challenge report.
\newblock {\em Computer Methods and Programs in Biomedicine}, 212:106452, 2021.

\bibitem{hughes2021krippendorffsalpha}
John Hughes.
\newblock krippendorffsalpha: An r package for measuring agreement using
  krippendorff's alpha coefficient.
\newblock {\em arXiv preprint arXiv:2103.12170}, 2021.

\bibitem{hutchinson2022analysis}
Kay Hutchinson, Zongyu Li, Leigh~A Cantrell, Noah~S Schenkman, and Homa
  Alemzadeh.
\newblock Analysis of executional and procedural errors in dry-lab robotic
  surgery experiments.
\newblock {\em The International Journal of Medical Robotics and Computer
  Assisted Surgery}, 18(3):e2375, 2022.

\bibitem{inouye2022assessing}
Daniel~A Inouye, Runzhuo Ma, Jessica~H Nguyen, Jasper Laca, Rafal Kocielnik,
  Anima Anandkumar, and Andrew~J Hung.
\newblock Assessing the efficacy of dissection gestures in robotic surgery.
\newblock {\em Journal of Robotic Surgery}, pages 1--7, 2022.

\bibitem{Kati2014model}
Darko Kati{\'{c}}, Anna-Laura Wekerle, Fabian G{\"a}rtner, Hannes Kenngott,
  Beat~Peter M{\"u}ller-Stich, R{\"u}diger Dillmann, and Stefanie Speidel.
\newblock Knowledge-driven formalization of laparoscopic surgeries for
  rule-based intraoperative context-aware assistance.
\newblock In Danail Stoyanov, D.~Louis Collins, Ichiro Sakuma, Purang
  Abolmaesumi, and Pierre Jannin, editors, {\em Information Processing in
  Computer-Assisted Interventions}, pages 158--167, Cham, 2014. Springer
  International Publishing.

\bibitem{Kitaguchi2021Artificial}
Daichi Kitaguchi, Nobuyoshi Takeshita, Hiro Hasegawa, and Masaaki Ito.
\newblock Artificial intelligence‐based computer vision in surgery: Recent
  advances and future perspectives.
\newblock {\em Annals of Gastroenterological Surgery}, 6, 10 2021.

\bibitem{lalys2014surgical}
Florent Lalys and Pierre Jannin.
\newblock Surgical process modelling: a review.
\newblock {\em International journal of computer assisted radiology and
  surgery}, 9(3):495--511, 2014.

\bibitem{lea2016temporal}
Colin Lea, Rene Vidal, Austin Reiter, and Gregory~D Hager.
\newblock Temporal convolutional networks: A unified approach to action
  segmentation.
\newblock In {\em European Conference on Computer Vision}, pages 47--54.
  Springer, 2016.

\bibitem{li2022sirnet}
Ling Li, Xiaojian Li, Shuai Ding, Zhao Fang, Mengya Xu, Hongliang Ren, and
  Shanlin Yang.
\newblock Sirnet: Fine-grained surgical interaction recognition.
\newblock {\em IEEE Robotics and Automation Letters}, 2022.

\bibitem{Li2022error}
Zongyu Li, Kay Hutchinson, and Homa Alemzadeh.
\newblock Runtime detection of executional errors in robot-assisted surgery.
\newblock In {\em 2022 International Conference on Robotics and Automation
  (ICRA)}, page 3850–3856. IEEE Press, 2022.

\bibitem{lin2010structure}
Henry~C Lin.
\newblock {\em Structure in surgical motion}.
\newblock The Johns Hopkins University, 2010.

\bibitem{ma2021novel}
Runzhuo Ma, Erik~B Vanstrum, Jessica~H Nguyen, Andrew Chen, Jian Chen, and
  Andrew~J Hung.
\newblock A novel dissection gesture classification to characterize robotic
  dissection technique for renal hilar dissection.
\newblock {\em The Journal of Urology}, 205(1):271--275, 2021.

\bibitem{meli2021unsupervised}
Daniele Meli and Paolo Fiorini.
\newblock Unsupervised identification of surgical robotic actions from small
  non homogeneous datasets.
\newblock {\em arXiv preprint arXiv:2105.08488}, 2021.

\bibitem{menegozzo2019surgical}
Giovanni Menegozzo, Diego Dall’Alba, Chiara Zandon{\`a}, and Paolo Fiorini.
\newblock Surgical gesture recognition with time delay neural network based on
  kinematic data.
\newblock In {\em 2019 International Symposium on Medical Robotics (ISMR)},
  pages 1--7. IEEE, 2019.

\bibitem{neumuth2011modeling}
Dayana Neumuth, Frank Loebe, Heinrich Herre, and Thomas Neumuth.
\newblock Modeling surgical processes: A four-level translational approach.
\newblock {\em Artificial intelligence in medicine}, 51(3):147--161, 2011.

\bibitem{nwoye2022cholectriplet2021}
Chinedu~Innocent Nwoye, Deepak Alapatt, Tong Yu, Armine Vardazaryan, Fangfang
  Xia, Zixuan Zhao, Tong Xia, Fucang Jia, Yuxuan Yang, Hao Wang, Derong Yu,
  Guoyan Zheng, Xiaotian Duan, Neil Getty, Ricardo Sanchez-Matilla, Maria Robu,
  Li~Zhang, Huabin Chen, Jiacheng Wang, Liansheng Wang, Bokai Zhang, Beerend
  Gerats, Sista Raviteja, Rachana Sathish, Rong Tao, Satoshi Kondo, Winnie
  Pang, Hongliang Ren, Julian~Ronald Abbing, Mohammad~Hasan Sarhan, Sebastian
  Bodenstedt, Nithya Bhasker, Bruno Oliveira, Helena~R. Torres, Li~Ling, Finn
  Gaida, Tobias Czempiel, João~L. Vilaça, Pedro Morais, Jaime Fonseca,
  Ruby~Mae Egging, Inge~Nicole Wijma, Chen Qian, Guibin Bian, Zhen Li,
  Velmurugan Balasubramanian, Debdoot Sheet, Imanol Luengo, Yuanbo Zhu, Shuai
  Ding, Jakob-Anton Aschenbrenner, Nicolas~Elini van~der Kar, Mengya Xu,
  Mobarakol Islam, Lalithkumar Seenivasan, Alexander Jenke, Danail Stoyanov,
  Didier Mutter, Pietro Mascagni, Barbara Seeliger, Cristians Gonzalez, and
  Nicolas Padoy.
\newblock Cholectriplet2021: A benchmark challenge for surgical action triplet
  recognition.
\newblock {\em arXiv preprint arXiv:2204.04746}, 2022.

\bibitem{Nwoye2020recginter}
Chinedu~Innocent Nwoye, Cristians Gonzalez, Tong Yu, Pietro Mascagni, Didier
  Mutter, Jacques Marescaux, and Nicolas Padoy.
\newblock Recognition of instrument-tissue interactions in endoscopic videos
  via action triplets.
\newblock In Anne~L. Martel, Purang Abolmaesumi, Danail Stoyanov, Diana Mateus,
  Maria~A. Zuluaga, S.~Kevin Zhou, Daniel Racoceanu, and Leo Joskowicz,
  editors, {\em Medical Image Computing and Computer Assisted Intervention --
  MICCAI 2020}, pages 364--374, Cham, 2020. Springer International Publishing.

\bibitem{nwoye2022rendezvous}
Chinedu~Innocent Nwoye, Tong Yu, Cristians Gonzalez, Barbara Seeliger, Pietro
  Mascagni, Didier Mutter, Jacques Marescaux, and Nicolas Padoy.
\newblock Rendezvous: Attention mechanisms for the recognition of surgical
  action triplets in endoscopic videos.
\newblock {\em Medical Image Analysis}, 78:102433, 2022.

\bibitem{park2012crowdsourcing}
Sunghyun Park, Gelareh Mohammadi, Ron Artstein, and Louis-Philippe Morency.
\newblock Crowdsourcing micro-level multimedia annotations: The challenges of
  evaluation and interface.
\newblock In {\em Proceedings of the ACM multimedia 2012 workshop on
  Crowdsourcing for multimedia}, pages 29--34, 2012.

\bibitem{qin2020davincinet}
Yidan Qin, Seyedshams Feyzabadi, Max Allan, Joel~W Burdick, and Mahdi Azizian.
\newblock davincinet: Joint prediction of motion and surgical state in
  robot-assisted surgery.
\newblock {\em arXiv preprint arXiv:2009.11937}, 2020.

\bibitem{qin2020temporal}
Yidan Qin, Sahba~Aghajani Pedram, Seyedshams Feyzabadi, Max Allan, A~Jonathan
  McLeod, Joel~W Burdick, and Mahdi Azizian.
\newblock Temporal segmentation of surgical sub-tasks through deep learning
  with multiple data sources.
\newblock In {\em 2020 IEEE International Conference on Robotics and Automation
  (ICRA)}, pages 371--377. IEEE, 2020.

\bibitem{rivas2021surgical}
Irene Rivas-Blanco, Carlos~J P{\'e}rez-del Pulgar, Andrea Mariani, Claudio
  Quaglia, Giuseppe Tortora, Arianna Menciassi, and V{\'\i}ctor~F Mu{\~n}oz.
\newblock A surgical dataset from the da vinci research kit for task automation
  and recognition.
\newblock {\em arXiv preprint arXiv:2102.03643}, 2021.

\bibitem{singh2021saras}
Vivek Singh~Bawa, Gurkirt Singh, Francis KapingA, Inna Skarga-Bandurova,
  Elettra Oleari, Alice Leporini, Carmela Landolfo, Pengfei Zhao, Xi~Xiang,
  Gongning Luo, Kuanquan Wang, Liangzhi Li, Bowen Wang, Shang Zhao, Li~Li,
  Armando Stabile, Francesco Setti, Riccardo Muradore, and Fabio Cuzzolin.
\newblock The saras endoscopic surgeon action detection (esad) dataset:
  Challenges and methods.
\newblock {\em arXiv e-prints}, pages arXiv--2104, 2021.

\bibitem{sutherland2015robotics}
Garnette~R Sutherland, Yaser Maddahi, Liu~Shi Gan, Sanju Lama, and Kourosh
  Zareinia.
\newblock Robotics in the neurosurgical treatment of glioma.
\newblock {\em Surgical Neurology International}, 6(Suppl 1):S1, 2015.

\bibitem{tao2012sparse}
Lingling Tao, Ehsan Elhamifar, Sanjeev Khudanpur, Gregory~D Hager, and Ren{\'e}
  Vidal.
\newblock Sparse hidden markov models for surgical gesture classification and
  skill evaluation.
\newblock In {\em International conference on information processing in
  computer-assisted interventions}, pages 167--177. Springer, 2012.

\bibitem{twinanda2016endonet}
Andru~P Twinanda, Sherif Shehata, Didier Mutter, Jacques Marescaux, Michel
  De~Mathelin, and Nicolas Padoy.
\newblock Endonet: a deep architecture for recognition tasks on laparoscopic
  videos.
\newblock {\em IEEE transactions on medical imaging}, 36(1):86--97, 2016.

\bibitem{valderrama2022towards}
Natalia Valderrama, Paola Ruiz~Puentes, Isabela Hern{\'a}ndez, Nicol{\'a}s
  Ayobi, Mathilde Verlyck, Jessica Santander, Juan Caicedo, Nicol{\'a}s
  Fern{\'a}ndez, and Pablo Arbel{\'a}ez.
\newblock Towards holistic surgical scene understanding.
\newblock In {\em International Conference on Medical Image Computing and
  Computer-Assisted Intervention}, pages 442--452. Springer, 2022.

\bibitem{van2021gesture}
Beatrice van Amsterdam, Matthew Clarkson, and Danail Stoyanov.
\newblock Gesture recognition in robotic surgery: a review.
\newblock {\em IEEE Transactions on Biomedical Engineering}, 2021.

\bibitem{van2020multi}
Beatrice van Amsterdam, Matthew~J Clarkson, and Danail Stoyanov.
\newblock Multi-task recurrent neural network for surgical gesture recognition
  and progress prediction.
\newblock In {\em 2020 IEEE International Conference on Robotics and Automation
  (ICRA)}, pages 1380--1386. IEEE, 2020.

\bibitem{varadarajan2009data}
Balakrishnan Varadarajan, Carol Reiley, Henry Lin, Sanjeev Khudanpur, and
  Gregory Hager.
\newblock Data-derived models for segmentation with application to surgical
  assessment and training.
\newblock In {\em International Conference on Medical Image Computing and
  Computer-Assisted Intervention}, pages 426--434. Springer, 2009.

\bibitem{wagner2021comparative}
Martin Wagner, Beat-Peter M{\"u}ller-Stich, Anna Kisilenko, Duc Tran, Patrick
  Heger, Lars M{\"u}ndermann, David~M Lubotsky, Benjamin M{\"u}ller, Tornike
  Davitashvili, Manuela Capek, Annika Reinke, Tong Yu, Armine Vardazaryan,
  Chinedu~Innocent Nwoye, Nicolas Padoy, Xinyang Liu, Eung-Joo Lee, Constantin
  Disch, Hans Meine, Tong Xia, Fucang Jia, Satoshi Kondo, Wolfgang Reiter,
  Yueming Jin, Yonghao Long, Meirui Jiang, Qi~Dou, Pheng~Ann Heng, Isabell
  Twick, Kadir Kirtac, Enes Hosgor, Jon~Lindström Bolmgren, Michael Stenzel,
  Björn~von Siemens, Hannes~G. Kenngott, Felix Nickel, Moritz~von Frankenberg,
  Lena Mathis-Ullrich, Franziska ad Maier-Hein, Stefanie Speidel, and Sebastian
  Bodenstedt.
\newblock Comparative validation of machine learning algorithms for surgical
  workflow and skill analysis with the heichole benchmark.
\newblock {\em arXiv preprint arXiv:2109.14956}, 2021.

\bibitem{ward2020training}
Thomas~M Ward, Daniel Hashimoto, Yutong Ban, Elan~R Witkowski, Keith~D
  Lillemoe, Guy Rosman, and Ozanan~R Meireles.
\newblock Training with pooled annotations from multiple surgeons has no effect
  on a deep learning artificial intelligence model’s performance.
\newblock {\em J Am Coll Surg}, 231(4):e203, 2020.

\bibitem{Xu2021report}
Mengya Xu, Mobarakol Islam, Chwee Ming~Lim, and Hongliang Ren.
\newblock Learning domain adaptation with model calibration for surgical report
  generation in robotic surgery.
\newblock In {\em 2021 IEEE International Conference on Robotics and Automation
  (ICRA)}, pages 12350--12356, 2021.

\bibitem{yasar2020real}
Mohammad~Samin Yasar and Homa Alemzadeh.
\newblock Real-time context-aware detection of unsafe events in robot-assisted
  surgery.
\newblock In {\em 2020 50th Annual IEEE/IFIP International Conference on
  Dependable Systems and Networks (DSN)}, pages 385--397. IEEE, 2020.

\bibitem{yasar2019context}
Mohammad~Samin Yasar, David Evans, and Homa Alemzadeh.
\newblock Context-aware monitoring in robotic surgery.
\newblock In {\em 2019 International Symposium on Medical Robotics (ISMR)},
  pages 1--7. IEEE, 2019.

\end{thebibliography}


\end{document}